\documentclass[10pt,twocolumn,letterpaper]{article}

\usepackage{cvpr}              

\usepackage{ulem}

\definecolor{cvprblue}{rgb}{0.21,0.49,0.74}
\usepackage[pagebackref,breaklinks,colorlinks,allcolors=cvprblue]{hyperref}
\usepackage{microtype}      
\usepackage[table]{xcolor}
\definecolor{darkgrey}{rgb}{0.4,0.4,0.4} 
\usepackage{graphicx}

\newcommand{\mv}[0]{\ensuremath{\boldsymbol{m}} }

\newcommand{\tv}[0]{\ensuremath{\boldsymbol{t}} }

\newcommand{\xv}[0]{\ensuremath{\boldsymbol{x}} }
\newcommand{\yv}[0]{\ensuremath{\boldsymbol{y}} }

\usepackage{enumitem}
\usepackage{multirow} 
\usepackage{caption}
\usepackage{tabularx}
\usepackage[table]{xcolor}
\usepackage{makecell}  

\usepackage{placeins}
\usepackage{ulem}

\definecolor{ccr}{RGB}{10,110,150}  

\usepackage{amsmath}
\usepackage{pifont}
\usepackage{graphicx}   %
\usepackage{algorithm}
\usepackage{algorithmic}

\usepackage[accsupp]{axessibility}  

\def\rr{\textcolor{red}}

\def\pk{\textcolor{magenta}}
\def\sky{\textcolor{cyan}}

\def\paperID{} 
\def\confName{CVPR}
\def\confYear{2026}

\title{%
  STiTch: Semantic Transition and Transportation in Collaboration for \\
  Training-Free Zero-Shot Composed Image Retrieval%
}

\author{
Miaoge Li$^{1\dag}$
,
Dongsheng Wang$^{2*}$
,
Zening Sun$^{2\dag}$
,
Jinsen Zhang$^2$
,
Wenhan Luo$^3$
,
Jingcai Guo$^{1}$\thanks{Corresponding authors. \dag Equal contribution.}
\\
$^1$Department of COMP/LSGI, The Hong Kong Polytechnic University, Hong Kong SAR\\
$^2$College of Computer Science and Software Engineering, Shenzhen University, Shenzhen, China\\
$^3$The Hong Kong University of Science and Technology, Hong Kong SAR\\
\texttt{jc-jingcai.guo@polyu.edu.hk}
}

\newcounter{appendixcnt}

\begin{document}
\maketitle
\begin{abstract}
Training-free zero-shot composed image retrieval models are recently gaining increasing research interest due to their generalizability and flexibility in unseen multimodal retrieval. Recent LLM-based advances focus on generating the expected target caption by exploring the compositional ability behind the LLMs. Although efficient, we find that 1) the generated captions tend to introduce unexpected features from the reference image due to the semantic gap between the input image and text modification, where the image contains much more details than the text; 2) the point-to-point alignment during the retrieval stage fails to capture diverse compositions.
To address these challenges, we introduce a novel \underline{\textbf{S}}emantic \underline{\textbf{T}}rans\underline{\textbf{i}}tion and \underline{\textbf{T}}ransportation in \underline{\textbf{c}}ollaboration framework for training-free zero-s\underline{\textbf{h}}ot CIR tasks. Specifically, given the composed caption inferred by an LLM, we aim to refine it through a transition vector in the embedding space and make it closer to the target image. Combining LLMs with user instruction, the refined caption concentrates more on the core modification intent and thus filters out unnecessary noise. Moreover, to explore diverse alignment during the retrieval stage, we model the caption and image as discrete distributions and reformulate the retrieval task as a set-to-set alignment task. Finally, a bidirectional transportation distance is developed to consider fine-grained alignments across modalities and calculate the retrieval score.
Extensive experiments
demonstrate that our method can be general, effective, and beneficial for many CIR tasks.
\end{abstract}

\section{Introduction}
\label{sec:intro}

Composed Image Retrieval (CIR) aims to search for a target image using a compositional query of a reference image and text modification~\cite{vo2019composing,lee2021cosmo,hosseinzadeh2020composed,chen2020image,baldrati2022effective}. 
One of the key challenges is to model the multimodal relationship of the triplet: $<$\textit{reference image, text modification, target image}$>$. Previous studies have focused on fusing the input image and modifications within a shared embedding space in a supervised manner~\cite{vo2019composing,delmas2022artemis,anwaar2021compositional}. Generally, these models typically rely on expensive manually-annotated triplets and often exhibit suboptimal performance in unseen scenarios~\cite{baldrati2023zero,karthik2023vision}. 
Motivated by the success of textual inversion in image generation~\cite{gal2022image,mokady2023null,wang2024instruction}, recent studies have proposed Zero-Shot Composed Image Retrieval (ZS-CIR)~\cite{saito2023pic2word,zeng2022socratic,HyCIR}. These models focus on training a mapping network to convert the reference image into continuous textual embeddings and then merge them with text modifications using static templates for target captions, enabling CIR without explicit supervision. Unfortunately, these models also need image-caption pairs to learn the mapping parameters, and the mismatch between textual inversion and CIR may hamper their ability to accurately infer the implicit user intent conveyed in the text modification.

\begin{figure*}[!t]
\centering
\includegraphics[width=0.9\textwidth]{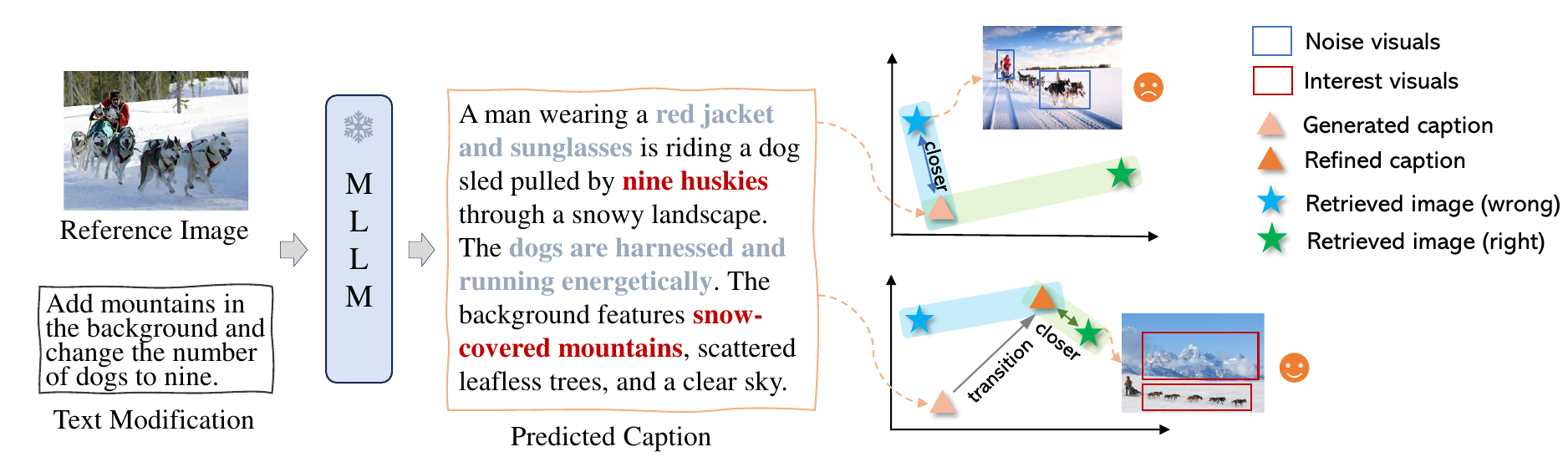}
\caption{\small{Motivation of our proposed model. Predicted captions from MLLMs typically consist of expected ground-truth sentences (red words) and unexpected visual details (gray words).}}
\label{motivation_1}
\vspace{-3mm}
\end{figure*}

Alternatively, training-free approaches paired with foundation models can achieve effective CIR without additional training and offer improved reasoning capabilities \cite{karthik2023vision,tang2025reason}. 
There are mainly two directions: two-stage methods typically require an image captioner and an LLM to first generate detailed captions of the reference image and then fuse them with text modification via an LLM to produce the target descriptions; one-stage methods unify this process by employing an MLLM to directly output the target captions given the multimodal queries.
Despite considerable progress, several challenges remain. First, the above generation-then-retrieval pipeline is prone to \textit{Extraneous Cognitive Load} \cite{sweller1988cognitive}.  
Specifically, the reference image may trigger information leakage, which in turn leads to overemphasis on irrelevant details, affecting the retrieval performance. As shown in Fig.\ref{motivation_1}, the target caption generated by MLLMs includes extraneous elements such as ``red jacket and sunglasses'', which are unrelated to the textual modification. Therefore, distracting from the core intent,  it diminishes the ability to identify key information, such as ``snow-covered mountains''. 
Second, most existing models generate either a single description or simply average multiple descriptions to obtain the final representation~\cite{tang2025reason,yang2024semantic}. However, as one image is worth a thousand words, such point-to-point alignment focuses on partial features and fails to capture complex relations. This inherent heterogeneity between visual and textual representations inevitably leads to semantic imbalance across modalities, leading to suboptimal retrieval prediction~\cite{zhu2024awt,chen2023plot,wang2023tuning}. 

To address the above issues, we proposes STiTch, a novel one-stage, training-free framework that improves the existing generation-then-retrieval pipeline by introducing \underline{\textbf{S}}emantic \underline{\textbf{T}}rans\underline{\textbf{i}}tion and \underline{\textbf{T}}ransportation in \underline{\textbf{c}}ollaboration framework for zero-s\underline{\textbf{h}}ot CIR tasks. Like previous works~\cite{tang2025reason}, we explore the in-context learning of MLLMs and directly query an MLLM to generate the target caption given the reference image and text modification. Importantly, to address the above asymmetry issues, STiTch utilizes the uncertainty ability of the language decoder and views the description as a discrete distribution by generating multiple candidates. Each candidate in the distribution focuses on a specific composition pattern, and they together provide a comprehensive understanding of the given query input.

Since reference images may inevitably introduce irrelevant information into captions generated by MLLMs, we propose guiding the textual caption toward the target image via a transition vector in the embedding space (as seen in Fig.~\ref{motivation_1}). Intuitively, an ideal transition vector should bridge the semantic gap between the generated caption and the target image. Here, we aim to solve it in a training-free manner and estimate the transition vector by feeding the text modification into the CLIP text encoder.
For one thing, since both the text modification and the generated caption share the same modality, the former can seamlessly refine the latter without introducing a modality gap or requiring extra parameters. For another, the text modification encapsulates the incremental, high-quality, and dense information that shifts from the reference image to the target image, guiding the model to refocus on core semantic information. This transition operates directly in the embedding space—simple yet efficient—and ensures that the final target caption retains diversity while reducing distortion, all of which is highly relevant and beneficial to the retrieval process. 

After obtaining high-quality and diverse features of target captions, it is crucial to align them more effectively with the target images in the embedding space. Similar to the textual domain, STiTch models each target image as a discrete distribution through multiple augmentations, where each captures local visuals of the target image. Together, these augmentations provide a rich visual representation from the visual domain, facilitating fine-grained alignments in the retrieval process.
Finally, a novel bidirectional transportation distance is further developed to calculate the similarity of two discrete distributions across the vision-language modalities. Specifically, given the cost matrix that measures the transport cost between the captions and image augmentation, STiTch designs both a forward path and a backward path to calculate the transportation distance from the caption set to the target image set and that from the target image set back to the caption set, respectively. This formulation effectively uncovers fine-grained cross-modal correlations by minimizing the bidirectional transport cost between modalities.

 The main contributions can be summarized as follows:

\begin{itemize}
    \item We propose a novel one-stage, training-free framework that considers both semantic transition and transportation collaboratively for ZS-CIR. By introducing the modification-driven transition to the generated caption, the feature drift caused by additional irrelevant noise is compensated in the embedding space.
    \item We elegantly transform the retrieval procedure into a bidirectional transport problem. Explicitly explore fine-grained alignments between diverse refined textual captions and enhanced target images.
    \item Extensive comparisons and ablations on four benchmarks demonstrate the effectiveness of the proposed STiTch with competitive performance in all settings.
\end{itemize}

\section{Related Work}
\label{sec:related_work}

\subsection{Composed Image Retrieval}
Composed Image Retrieval (CIR) has inspired various architectural innovations \cite{vo2019composing,chen2020image,lee2021cosmo}. Early methods adopt a fusion paradigm to learn joint embeddings of reference image and modification features via contrastive or attention-based objectives \cite{chen2020learning,anwaar2021compositional}. Some recent works train large-scale retrieval models on millions of web-mined triplets\cite{zhang2024magiclens}, but reliance on supervision limits scalability. To address this, Zero-Shot CIR (ZS-CIR) enables retrieval without labeled data \cite{tang2025missing}. For example, Pic2Word \cite{anwaar2021compositional} projects image features into a token embedding space, while SEARLE \cite{baldrati2023zero} improves alignment via a text inversion network. Another line of work generates synthetic triplets from image-caption pairs using generative models \cite{gu2023compodiff}.

With the advent of foundation models, recent studies have tackled CIR in a training-free manner, leveraging their
strong contextual understanding. 
Two main paradigms have emerged: two-stage methods performs reference image captioning and text manipulation separately, while one-stage methods generate target captions directly from multimodal inputs.
For example,
CIReVL\cite{karthik2023vision} initially employs pre-trained captioning models
\cite{li2023blip} to generate caption for a given image. Subsequently, it queries an LLM to refine and recompose the caption based on text modifications for text-to-image retrieval. LDRE\cite{yang2024ldre} considers diverse semantics of the CIR and generates diverse captions at the first stage, and then adopts an ensemble strategy to get the final single feature for the multiple captions.
OSrCIR\cite{tang2025reason}
uses MLLMs to infer user intent by directly processing a query pair, guided by a reflective chain-of-thought prompt.  
However, both are overwhelmed by the rich semantic information from the reference image, which may overshadow the key modifications. Different from existing LLM-based methods, our STiTch aims to refine the captions after the generation and improve semantic alignments by formulating the retrieval task as the bidirectional transportation problem.


\subsection{Alignment via Transport Distance}
Recently, Optimal Transport (OT) \cite{villani2009optimal} has been widely used for aligning distributions
in various domains \cite{redko2019optimal,zhao2018label,zhao2018label,lee2019hierarchical,chen2020graph,wang2023tuning,zhu2025dynamic}.
Unlike traditional distance metrics like Euclidean distance, 
OT provides a more geometrically nuanced measure that captures structural similarities between distributions.
However, it typically requires 
iterative optimization via the
Sinkhorn algorithm \cite{cuturi2013sinkhorn}, which can be time-consuming. To this end, Conditional Transport (CT) considers the transport plan based on the semantic similarity between samples from two distributions bidirectionally \cite{zheng2021exploiting}. Its flexibility allows seamless integration with deep learning frameworks, offering lower computational complexity
and better scalability, resulting in superior performance in recent alignment tasks \cite{liu2023patch,li2023patchct,li2024tsca}. For instance, \cite{tian2023prototypes} exploits transferable statistics with CT to refine biased prototypes to capture unbiased statistics within imbalanced query samples. \cite{li2023patchct} design a sparse and layer-wise CT framework to enhance interactions between visual patches and textual labels, ensuring higher semantic consistency for multi-label classification. 
Notably, CT is inherently well-suited for a key step in CIR—aligning multiple captions with target images—especially when multimodal representations are involved. Motivated by this potential, we transform the traditional point-to-point similarity measure into a minimization of CT-based distance from a distributional perspective.
\section{Method}
\label{sec:method}

\begin{figure*}[!ht]
\centering
\includegraphics[width=0.82\textwidth]{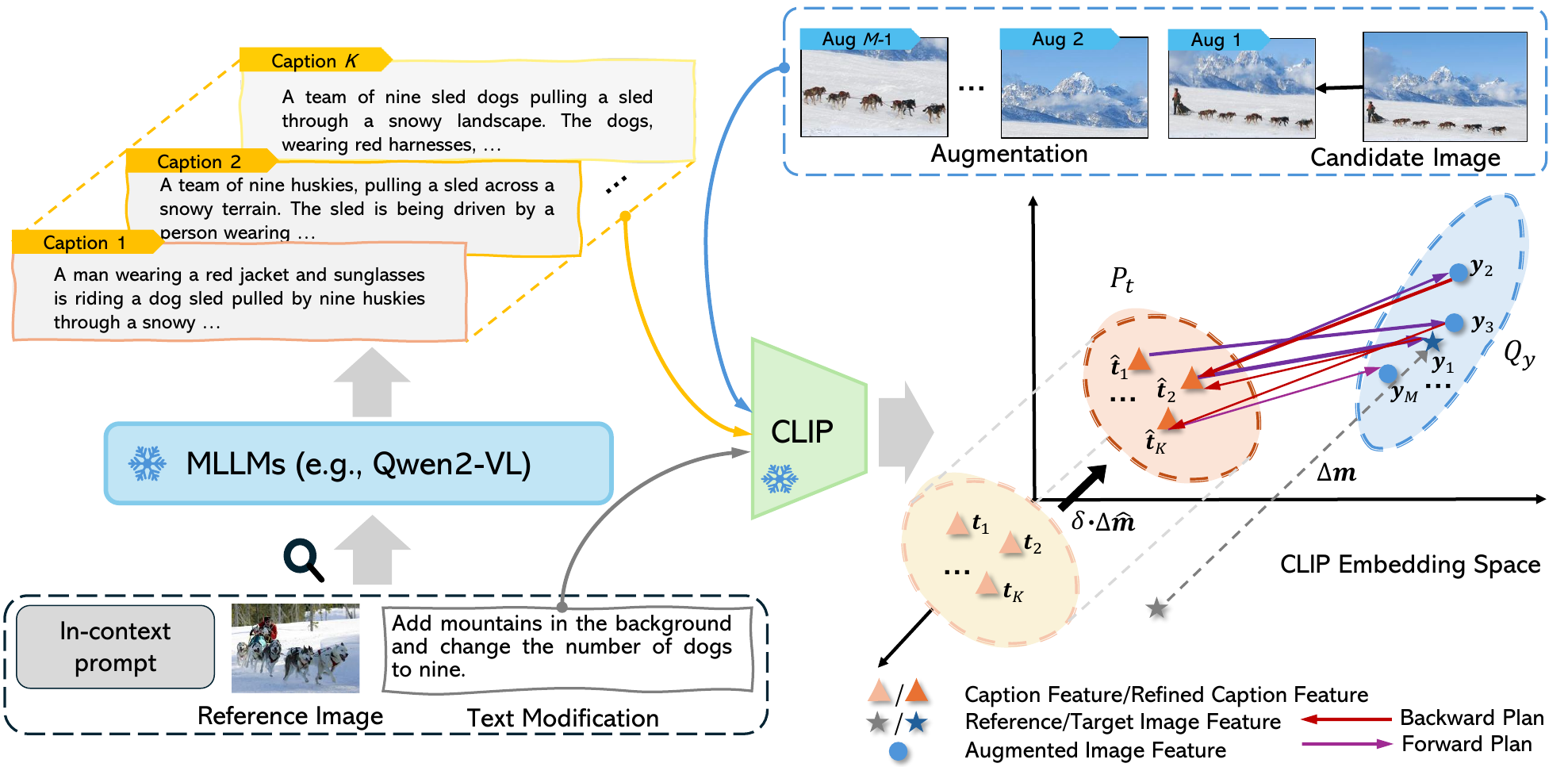}
\caption{\small{The overall framework of our method. STiTch first queries MLLMs to generate multiple captions and then refines them towards the target image via the transition vector in the embedding space. Subsequently, the refined captions and augmented target images are modeled as discrete distributions, enabling fine-grained set-to-set alignment via bidirectional transportation.}}
\vspace{-4mm}
\label{framework}
\end{figure*}

\noindent \textbf{Preliminaries.}
Let us denote $(x, m, y)$ as the CIR triplet \textit{$<$reference image, text modification, target image$>$}, respectively. The first two are multimodal inputs from users describing their retrieval intent. Training-free ZS-CIR aims to search a target image 
$y$
from an image database $\mathit{Y}=\{y_n\}_{n=1}^N$ that satisfies the semantic consistency with both $x$ and $m$, without requiring additional training. Generally, existing 
approaches
follow a generation-then-retrieval pipeline to make the final prediction. They first feed the reference image $x$ and text modification $m$ into a fusing model (such as an MLLM) to obtain the composed description of the target image, denoted as $t=\text{MLLM}(x, m)$. The retrieval score is then calculated by CLIP similarity:
\begin{equation}\label{clip}
    p(y=i|x,m) = \frac{\text{exp}(-dis(\tv, \yv_i) / \tau)}{\sum_{n=1}^N \text{exp}(-dis(\tv, \yv_n) / \tau)},
\end{equation}
where $\tv \in R^d$ and $\yv \in R^d$ are the latent features of $t$ and $y$ in CLIP space, with $d$ denotes the embedding dimension. $dis$ is the distance function and $\tau$ is the temperature parameter. $\tv$ in Eq.~\ref{clip} can be viewed as prototypes that capture reasonable visual features of $y$. To find the optimal $\tv$, recent LLM-based models develop various attempts, including two-stage generation and chain-of-thought reasoning\cite{yang2024ldre,li2025imagine,tang2025reason}. 

Despite their promising results, We find that 1) the generated description $\tv$ usually inherits the unnecessary details from the reference image, leading to suboptimal prototype learning; 2) The point estimation of $\tv$ fails to model complex composed relations, this may limit the uncertainty of $\tv$ and diminish the generalizability.

\subsection{{Semantic Transition and Transportation}}
Starting from Eq.~\ref{clip}, we propose a novel training-free ZS-CIR framework to solve the mentioned shortcomings, as
illustrated in Fig.~\ref{framework}. Specifically, our proposed model consists of three modules: Querying, Transition, and Alignment. 
Unlike previous point estimation of $\tv$, STiTch views the composed prototype as a discrete distribution, e.g., $P_t$, over the caption space. This allows $P_t$ to focus on various possible target captions, showing higher diversity. To alleviate the issues of unnecessary information pollution, we further transfer the obtained $\tv$ using the text modification $\mv$ in the embedding space. $\mv$ contains high-quality relative information linking the reference image to its target image. Thus, a simple combination strategy is developed to push $\tv$ to its target image, resulting in more precise $\tv$. Finally, we also view the target image as a discrete distribution and develop a bidirectional transportation distance to align the composed prototypes and target images for fine-grained retrieval. 

\noindent \textbf
{Querying.} As discussed above, one of the core challenges is to generate reasonable captions $\tv$ for image $y$. Inspired by previous works \cite{li2023blip,li2024improving}, STiTch aims to solve this with MLLMs due to their impressive performance in image-text understanding. More specifically, here we explore the in-context learning of MLLM and complete the prompt template: ``\textit{$<$in-context prompt$>$. $<x>$. Instruction:$<m>$. Edited Description:}''. Where \textit{$<$in-context prompt$>$} helps MLLMs understand the CIR task and output the expected target descriptions. $<x>$, $<m>$ are the placeholder of the reference image and text modification. 

Intuitively, there are likely several plausible $t$ for each $x$ and $m$ pair, they describe the same target image from different views, To simulate such an ability, we explore the uncertainty generation of MLLMs via sampling from the language model's decoder, to replace the naive greedy decoding used in previous single target description generations. Formally, we combine the top-k and top-p sampling strategy in ~\citep{holtzman2019curious} and collect $K$ possible target description $t$ with a discrete distribution:
\begin{equation}\label{P_t}
    P_t = \frac{1}{K}\sum_{k=1}^K \delta_{\tv_k},
\end{equation}
where $\delta_{\tv}$ refers to a point mass located at coordinate $\tv$, and $\tv_k$ denotes the text embedding of $k$-th generated description. $P_t$ can be viewed as a semantic set containing $K$ reasonable descriptions, and it thus considers diverse visual features of the target image. 

\vspace{2mm}
\noindent \textbf{Semantic Transition.} Another challenge comes from the semantic gap between the generated description $t$ and the target image $y$. Generally, an ideal $t$ should highlight the semantic changes while avoiding unnecessary information instructions from the reference image. 
On one hand, we empirically find that existing LLM-based generators can successfully describe the changed context. On the other hand, they also pay more attention to the reference image due to the limited guidance in text modification $m$. As a result, the output caption usually contains many visual details of the reference image, which act as noise and mislead the retrieval process. To this end, we propose a transition step that pushes the generated captions to the target image at the embedding space. 
As shown in Fig.~\ref{framework}, let $\Delta \mv$ denote the difference between $\xv$ and $\yv$, it provides incremental semantics from the reference image to its target. Recalling that the text modification $m$ contains high-quality relative information, it is natural to estimate $\Delta \mv$ using $m$:
\begin{equation}\label{Delta}
    \Delta \mv = \yv - \xv 
    , 
     \Delta \hat{\mv} = f(m) ,
\end{equation}
where $\xv \in R^d$ represents the visual embeddings of the reference image $x$ generated by the CLIP image encoder, $f$ is the CLIP text encoder. Once obtaining the relative guidance $\Delta \hat{\mv}$, $\tv_k$ can be updated via a simple fusing strategy:
\begin{equation}\label{P_tv2}
    \hat{\tv}_k = (1-\alpha) \tv_k + \alpha \Delta \hat{\mv},
\end{equation}
where the first term $\tv_k$ comes from the MLLMs, and it encodes multimodal knowledge based on the MLLM's understanding of the composed input $(x, m)$. The second term $\Delta \hat{\mv}$ is derived from the text modification estimation and contains high-quality relative instruction between the reference and target image. The transferred caption $\hat{\tv}_k$ takes guidance from both directions with a trade-off hyperparameter $\alpha \in [0,1]$. Now, we can rewrite $P_t$ as: $P_t = \frac{1}{K}\sum_{k=1}^K \delta_{\hat{\tv}_k}$.  

\vspace{2mm}
\noindent \textbf{Set-to-set Alignment.} Given the collected discrete distribution $P_t$ in the text domain, we in this section aim to explore the diverse visual features in the image domain with a similar motivation:
\begin{equation}\label{Q_y}
    Q_y = \frac{1}{M}\sum_{m=1}^M \delta_{{\yv}_m},
\end{equation}
where we augment the target image $M-1$ times and $\{\yv_m\}_{m=2}^{M}$ are the embeddings of the augmented images.
Unlike previous ZS-CIR models that view the target image as a single point, $Q_y$ in Eq.~\ref{Q_y} provides us with multiple views of $y$, leading to the following fine-grained retrieval strategy. 

Moving beyond the point-to-point similarity measurement in Eq.~\ref{clip}, we here develop a bidirectional distance of two discrete distributions $\mathcal{L}_{bi}(P_t, Q_y)$ under the 
CT
framework. Specifically, $\mathcal{L}_{bi}$ consists of two transport costs: the forward cost that measures the expected transport cost from the refined captions to the target image and the backward cost that inverses the direction:
\begin{equation}\label{lbi}
\begin{aligned}
    \mathcal{L}_{bi}(P_t, Q_y) &= \mathcal{L}_{P_t\rightarrow Q_y}(P_t, Q_y) + \mathcal{L}_{Q_y \rightarrow P_t}(P_t, Q_y) \\
    &= \sum_{m,k} \pi(\yv_m|\hat{\tv}_k)c(\hat{\tv}_k,\yv_m) + \pi(\hat{\tv}_k|\yv_m)c(\yv_m,\hat{\tv}_k),
\end{aligned}
\end{equation}
where the cost function $c(\yv,\hat{\tv})=c(\hat{\tv},\yv)$ is specified as the cosine distance to measure the transport cost between points $\hat{\tv}$ and $\yv$. $\pi(\yv|\hat{\tv})$ denotes the transport plan in the forward path, and it measures how likely $\hat{\tv}$ will be transported to $\yv$:
\begin{equation}
    \pi(\yv_m|\hat{\tv}_k) = \frac{\text{exp}(\hat{\tv}_k^T\yv_m / \tau)}{\sum_{m'=1}^M \text{exp}(\hat{\tv}_k^T\yv_{m'} / \tau)}.
\end{equation}
Naturally, the closer $\hat{\tv}$ and $\yv$ are in the embedding space, the higher the transport probability from $\hat{\tv}$ to $\yv$. $\pi(\hat{\tv}|\yv)$ is defined in a similar way:
\begin{equation}
    \pi(\hat{\tv}_k|\yv_m) = \frac{\text{exp}(\yv_m^T\hat{\tv}_k / \tau)}{\sum_{k'=1}^K \text{exp}(\yv_m^T\hat{\tv}_{k'} / \tau)}.
\end{equation}
Mathematically, $\mathcal{L}_{bi}$ in Eq.~\ref{lbi} calculates the distance between two discrete distributions in both forward and backward directions.  This benefits the alignment across the vision and language domains, showing better multimodal retrieval ability. Moreover, $\mathcal{L}_{bi}$ views the generated caption and target image as two discrete distributions, which show great potential in modeling diverse semantics.

Once obtain the bidirectional distance between the generated descriptions and the target image, we can rewrite Eq.~\ref{clip} with $\mathcal{L}_{bi}$, resulting in a more general and fine-grained prediction score:
\begin{equation}\label{qta}
    p(y=i|x,m) = \frac{\text{exp}(-\mathcal{L}_{bi}(P_t, Q_{y_i}))}{\sum_{n=1}^N \text{exp}(-\mathcal{L}_{bi}(P_t, Q_{y_n})}.
\end{equation}
We summarize the whole inference algorithm of STiTch in the Alg.{\color{cvprblue}1}
in \pk{Appendix}.

\section{Experiments}
\label{sec:experiment}

\subsection{Experimental Setup}
\paragraph{Datasets.} Following previous works~\cite{tang2025reason}, we evaluate our proposed model on four commonly used CIR datasets, which vary in CIR tasks, image domains and dataset sizes, including CIRR~\cite{liu2021image}, CIRCO~\cite{baldrati2023zero}, FashionIQ~\cite{wu2021fashion}, and GeneCIS~\cite{vaze2023genecis}. CIRR is the first natural image dataset for CIR. CIRCO comes from COCO2017~\cite{lin2014microsoft} and has multiple ground truths for each query. FashionIQ focuses on fashion-related retrieval and consists of three subsets: shirt, dress, and toptee. GeneCIS contains images from MS-COCO~\cite{lin2014microsoft} and Visual Attributes in the Wild~\cite{pham2021learning}, offering four task variations around objects and attributes. We report the original benchmark metrics for each dataset: \textit{e.g.}, Recall@k(R@k) for CIRR, GeneCIS, and FashionIQ, and mean average precision (mAP@k) for CIRCO due to its multiple labels. 

\vspace{2mm}
\noindent \textbf{Baselines.} We compare our STiTch with recent advances, grouped as training-dependent and training-free models. The former often optimize a mapping network to project the reference image into text tokens, including 1) \textbf{Pic2Word}~\cite{saito2023pic2word}, 2) \textbf{SEARLE}~\cite{baldrati2023zero}, 3) \textbf{Context-I2W}~\cite{tang2024context}, and 4) \textbf{LinCIR}~\cite{gu2023language}. Training-free methods focus more on improving ZS-CIR with large language models, including 5) \textbf{CIReVL}~\cite{karthik2023vision}, 6) \textbf{LDRE}~\cite{yang2024ldre}, 7) \textbf{OSrCIR}~\cite{tang2025reason} and 8) \textbf{SEIZE}~\cite{yang2024semantic}. Unlike previous training-free methods, our STiTch aims to refine the composed caption in the embedding space and explore fine-grained alignment across vision-language domains. For all training-based baselines, we directly report the results according to their official papers. For training-free models, to make a fair comparison, we reproduce their results on the FashionIQ dataset according to their released codes and report results on other datasets according to the original papers. We encourage readers to \pk{Appendix} for more details.

\begin{table*}[!t]
\centering
\caption{\small{Comparison on CIRCO and CIRR datasets. The top two training-free results are highlighted in \textbf{bold} and \underline{underline}, respectively. 
}}
\label{circo_table}
\resizebox{0.9\textwidth}{!}{%
\begin{tabular}{c|c|c|cccc||ccc|ccc}
\toprule
\multicolumn{3}{c}{\textbf{CIRCO + CIRR $\rightarrow$}} & \multicolumn{4}{c}{\textbf{CIRCO}} & \multicolumn{6}{c}{\textbf{CIRR}} \\
\hline
\multicolumn{3}{c}{Metric} & \multicolumn{4}{c}{mAP@k} & \multicolumn{3}{c}{Recall@k} & \multicolumn{3}{c}{$\text{Recall}_{\text{Subset}}$@k} \\
Arch & Method & Train & k=5 & k=10 & k=25 & k=50 & k=1 & k=5 & k=10 & k=1 & k=2 & k=3 \\
\hline
\multirow{7}{*}{ViT-L/14}  
& Pic2Word & \textcolor{red}{\ding{51}} & 8.72 & 9.51 & 10.64 & 11.29 & 23.90 & 51.70 & 65.30 & 53.76 & 74.46 & 87.08 \\
& SEARLE & \textcolor{red}{\ding{51}} & 11.68 & 12.73 & 14.33 & 15.12 & 24.24 & 52.48 & 66.29 & 53.76 & 75.01 & 88.19 \\
& LinCIR & \textcolor{red}{\ding{51}} & 12.59 & 13.58 & 15.00 & 15.85 & 25.04 & 53.25 & 66.68 & 57.11 & 77.37 & 88.89 \\
& Context-I2W & \textcolor{red}{\ding{51}} & 13.04 & 14.62 & 16.14 & 17.16 & 25.60 & 55.10 & 68.50 & - & - & - \\
& CIReVL & \textcolor{green}{\ding{55}} & 18.57 & 19.01 & 20.89 & 21.80 & 24.55 & 52.31 & 64.92 & 59.54 & 79.88 & 89.69    \\
& LDRE & \textcolor{green}{\ding{55}} & 23.35 & 24.03 & 26.44 & 27.50 & 26.53 & 55.57 & 67.54 & 60.43 & 80.31 & 89.90 \\
& OSrCIR & \textcolor{green}{\ding{55}} & {23.87} & {25.33} & {27.84} & \underline{28.97} & \textbf{29.45} & \underline{57.68} & \underline{69.86} & {62.12} & {81.92} & {91.10} \\
& SEIZE & \textcolor{green}{\ding{55}} & \underline{24.98} & \underline{25.82} & \underline{28.24} & 28.35 & 28.65 & 57.16 & 69.23 & \underline{62.22} & \underline{84.05} & \underline{92.34} \\
& \cellcolor{gray!20} \textbf{STiTch(Ours)} & \cellcolor{gray!20}\textcolor{green}{\ding{55}} &  \cellcolor{gray!20}\textbf{25.55} & \cellcolor{gray!20}\textbf{26.27} & \cellcolor{gray!20}\textbf{28.81} & \cellcolor{gray!20}\textbf{29.99} & \cellcolor{gray!20}\underline{28.87} & \cellcolor{gray!20}\textbf{57.97} & \cellcolor{gray!20}\textbf{69.90} & \cellcolor{gray!20}\textbf{65.22} & \cellcolor{gray!20}\textbf{84.10} & \cellcolor{gray!20}\textbf{92.37}\\
\midrule
\multirow{3}{*}{ViT-G/14}
& CIReVL & \textcolor{green}{\ding{55}} & 26.77 & 27.59 & 29.96 & 31.03 & 34.65 & 64.29 & 75.06 & 67.95 & 84.87 & 93.21 \\
& LDRE & \textcolor{green}{\ding{55}} & {31.12} & {32.24} & 34.95 & 36.03 & 36.15 & 66.39 & 77.25 & 68.82 & {85.66} & {93.76} \\
& OSrCIR & \textcolor{green}{\ding{55}} & 30.47 & 31.14 & {35.03} & {36.59} & {37.26} & {67.25} & {77.33} & {69.22} & 85.28 & 93.55 \\
& SEIZE & \textcolor{green}{\ding{55}} & \underline{32.46} & \underline{33.77} & \underline{36.46} & \underline{37.55} & \underline{38.87} & \underline{69.42} & \underline{79.42} & \textbf{74.15} & \underline{89.23} & \underline{95.71} \\
& \cellcolor{gray!20} \textbf{STiTch (Ours)} & \cellcolor{gray!20}\textcolor{green}{\ding{55}} & \cellcolor{gray!20}\textbf{34.40} & \cellcolor{gray!20}\textbf{35.56} & \cellcolor{gray!20}\textbf{38.07} & \cellcolor{gray!20}\textbf{40.02} & \cellcolor{gray!20}\textbf{39.23} & \cellcolor{gray!20}\textbf{69.95} & \cellcolor{gray!20}\textbf{79.56} & \cellcolor{gray!20}\underline{73.56} & \cellcolor{gray!20}\textbf{89.50} & \cellcolor{gray!20}\textbf{95.86} \\
\bottomrule
\end{tabular}}
\vspace{-3mm}
\end{table*}

\vspace{2mm}
\noindent \textbf{Implementation Details.} We employ the open-source Qwen2-VL-7B as our MLLM by default, while we also report the results on various MLLMs in {\pk{Appendix} 
{\color{cvprblue}C}
}. To generate diverse descriptions, we follow a similar setting to previous works and apply $\tau=0.7$, top-k(k=50), top-p(p=0.8) at the querying stage. Regarding image augmentation, we used only random resized crop and random horizontal flip for each image. We set the number of captions as $K=5$, the number of augmentations as $M=10$, and employ a default value of $\alpha=0.45$. The default retrieval model is CLIP-L/14 from the official OpenAI implementation~\cite{radford2021learning}. We also report the results of LLM-based models (CIReVL, OSrCIR, and our STiTch) on CLIP-bigG-14 from the OpenCLIP implementation~\cite{cherti2023reproducible} for analysis of scaling laws.
All experiments are conducted on a single NVIDIA A6000 GPU.

\subsection{Comparison with State-of-the-art Methods}
We run all experiments three times with different random seeds and report the mean value to ensure reliability. We report the comparison on the hidden test set of CIRCO and CIRR datasets in Tab.~\ref{circo_table} (All results are obtained from the submission server provided in \cite{baldrati2023zero} and \cite{liu2021image}). These two CIR datasets focus on foreground and background differentiation and fine-grained image editing. From the results, we find that our proposed STiTch achieves the best or second-best results in most cases among baselines, including training-free and textual inversion models. STiTch underperforms baselines on the CIRR dataset in the case of $k=1$. This may be due to the noisy annotation in CIRR, where the reference image is less related to the target image~\cite{baldrati2023zero}. 

\begin{table*}[!t]
\centering
\caption{\small{Comparison on Fashion-IQ datasets. The top two training-free results are highlighted in \textbf{bolded} and \underline{underlined}. 
}}
\vspace{-2mm}
\label{fashioniq}
\resizebox{0.8\textwidth}{!}{%
\begin{tabular}{c|c|c|cccccc|cc}
\toprule
\multicolumn{3}{c}{\textbf{Fashion-IQ $\rightarrow$}} & \multicolumn{2}{c}{\textbf{Shirt}} & \multicolumn{2}{c}{\textbf{Dress}} & \multicolumn{2}{c}{\textbf{Toptee}} & \multicolumn{2}{c}{\textbf{Average}} \\
\hline
Arch & Method & Train & R@10 & R@50 & R@10 & R@50 & R@10 & R@50 & R@10 & R@50 \\
\hline
\multirow{6}{*}{ViT-G/14} 
& Pic2Word & \textcolor{red}{\ding{51}} & 33.17 & 50.39 & 25.43 & 47.65 & 35.24 & 57.62 & 31.28 & 51.89 \\
& SEARLE & \textcolor{red}{\ding{51}} & 36.46 & 55.35 & 28.16 & 50.32 & 39.83 & 61.45 & 34.81 & 55.71 \\
& LinCIR & \textcolor{red}{\ding{51}} & {46.76} & {65.11} & {38.08} & {60.88} & {50.48} & {71.09} & {45.11} & {65.69} \\
& CIReVL & \textcolor{green}{\ding{55}} &  29.85 & 51.07 & 27.07 & 49.53 & 35.80 & 56.14 & 32.19 & 52.36 \\
& OSrCIR & \textcolor{green}{\ding{55}} & {{38.65}} & {54.71} & {33.02} & {54.78} & {41.04} & {61.83} & {37.57} & {57.11} \\
& SEIZE & \textcolor{green}{\ding{55}} & \textbf{43.60} & \textbf{65.42} & \textbf{39.61} & \textbf{61.02} & \textbf{45.94} & \textbf{71.12} & \textbf{43.05} & \textbf{65.85} \\
& \cellcolor{gray!20}\textbf{STiTch(Ours)} & \cellcolor{gray!20}\textcolor{green}{\ding{55}} & \cellcolor{gray!20}{\underline{39.48}}  & \cellcolor{gray!20}{\underline{56.59}} & \cellcolor{gray!20}{\underline{35.04}} & \cellcolor{gray!20}{\underline{56.74}} & \cellcolor{gray!20}{\underline{42.86}} & \cellcolor{gray!20}{\underline{64.95}} & \cellcolor{gray!20}{\underline{39.12}} &  \cellcolor{gray!20}{\underline{59.43}} \\
\bottomrule
\end{tabular}}
\end{table*}

\begin{table*}[!t]
\centering
\caption{\small{Comparison on GeneCIS datasets. The top two training-free results are highlighted in \textbf{bolded} and \underline{underlined}, respectively.}}
\vspace{-2mm}
\label{genecis}
\resizebox{\textwidth}{!}{%
\begin{tabular}{c|c|c|cccccccccccc|c}
\toprule
\multicolumn{3}{c}{\textbf{GeneCIS $\rightarrow$}} & \multicolumn{3}{c}{\textbf{Focus Attribute}} & \multicolumn{3}{c}{\textbf{Change Attribute}} & \multicolumn{3}{c}{\textbf{Focus Object}} & \multicolumn{3}{c}{\textbf{Change Object}} & \multicolumn{1}{c}{\textbf{Average}}  \\
\hline
Arch & Method & Train & R@1 & R@2 & R@3 & R@1 & R@2 & R@3 & R@1 & R@2 & R@3 & R@1 & R@2 & R@3 & R@1 \\
\hline
\multirow{7}{*}{ViT-L/14} 
& SEARLE & \textcolor{red}{\ding{51}} & 17.1 & 29.6 & 40.7 & 16.3 & 25.2 & 34.2 & 12.0 & 22.2 & 30.9 & 12.0 & 24.1 & 33.9 & 14.4 \\
& LinCIR & \textcolor{red}{\ding{51}} & 16.9 & 30.0 & 41.5 & 16.2 & 28.0 & 36.8 & 8.3 & 17.4 & 26.2 & 7.4 & 15.7 & 25.0 & 12.2 \\
& Context-I2W & \textcolor{red}{\ding{51}} & 17.2 & 30.5 & 41.7 & 16.4 & 28.3 & 37.1 & 8.7 & 17.9 & 26.9 & 7.7 & 16.0 & 25.4 & 12.7\\
& CIReV & \textcolor{green}{\ding{55}} & 19.5 & 31.8 & 42.0 & 14.4 & 26.0 & 35.2 & 12.3 & 21.8 & 30.5 & 17.2 & 28.9 & 37.6 & 15.9\\
& OSrCIR & \textcolor{green}{\ding{55}} & \textbf{20.9} & {33.1} & {44.5} & {17.2} & {28.5} & {37.9} & {15.0} & {23.6} & {34.2} & {18.4} & {30.6} & {38.3} & {17.9} \\
& SEIZE & \textcolor{green}{\ding{55}} & \underline{20.5} & \underline{33.4} & \underline{45.0} & \underline{17.6} & \underline{28.9} & \underline{38.5} & \underline{15.4} & \underline{25.6} & \underline{36.2} & \underline{18.7} & \underline{30.9} & \underline{39.8} & \underline{18.1} \\
& \cellcolor{gray!20}\textbf{STiTch(Ours)} & \cellcolor{gray!20}\textcolor{green}{\ding{55}} &\cellcolor{gray!20}\uline{20.3} & \cellcolor{gray!20}\textbf{34.6} & \cellcolor{gray!20}\textbf{46.4} & \cellcolor{gray!20}\textbf{18.3} & \cellcolor{gray!20}\textbf{29.8} & \cellcolor{gray!20}\textbf{41.6} & \cellcolor{gray!20}\textbf{16.8} & \cellcolor{gray!20}\textbf{28.5} & \cellcolor{gray!20}\textbf{38.4} & \cellcolor{gray!20}\textbf{18.8} & \cellcolor{gray!20}\textbf{31.0} & \cellcolor{gray!20}\textbf{40.3} & \cellcolor{gray!20}\textbf{18.6} \\
\midrule
\multirow{3}{*}{ViT-G/14}
& CIReVL & \textcolor{green}{\ding{55}} & 20.9 & 34.4 & 44.9 & 16.5 & 29.0 & 39.8 & 15.1 & 25.6 & 33.4 & 18.5 & 31.6 & 41.4 & 17.8 \\
& OSrCIR & \textcolor{green}{\ding{55}} & \underline{22.7} & \underline{36.4} & 47.0 & 17.9 & 30.8 & 42.0 & 16.9 & 28.4 & 36.7 & \textbf{21.0} & \textbf{33.4} & \textbf{44.2} & 19.6 \\
& SEIZE & \textcolor{green}{\ding{55}} & \textbf{22.9} & 36.2 & \underline{47.3} & \underline{18.6} & \underline{31.4} & \underline{42.7} & \underline{18.2} & \underline{28.8} & \underline{37.6} & {19.6} & {33.0} & \underline{43.5} & \underline{19.8} \\
& \cellcolor{gray!20}\textbf{STiTch (Ours)} & \cellcolor{gray!20}\textcolor{green}{\ding{55}} &\cellcolor{gray!20}{21.9} & \cellcolor{gray!20}\textbf{36.4} &\cellcolor{gray!20}\textbf{47.9} &\cellcolor{gray!20}\textbf{19.6} &\cellcolor{gray!20}\textbf{31.9} &\cellcolor{gray!20}\textbf{42.8} &\cellcolor{gray!20}\textbf{20.2} &\cellcolor{gray!20}\textbf{30.3} &\cellcolor{gray!20}\textbf{39.6} & \cellcolor{gray!20}\underline{19.7} &\cellcolor{gray!20}\underline{33.2} &\cellcolor{gray!20}\uline{43.4} &\cellcolor{gray!20} \textbf{20.4} \\
\bottomrule
\end{tabular}}
\vspace{-2mm}
\end{table*}

Tab.~\ref{fashioniq} shows the comparison in the validation set of the Fashion-IQ dataset with ViT-G/14, and more comparisons on other backbones can be found in the \pk{Appendix} 
Tab. {\color{cvprblue}7}. The results show the robustness of our method. Interestingly, we find that text inversion-based models (e.g., LinCIR) outperform LLM-based models in the fashion image editing task. This may be due to the fact that most images in Fashion-IQ are relatively simple, containing only a pure background. This may limit the ability of MLLMs, and in contrast, training-dependent models tend to describe the reference image more correctly. 




Lastly, we further test the object and attribute composition ability of our model on the GeneCIS dataset, with the results listed in Tab.~\ref{genecis}. Unlike previous datasets that provide a detailed text modification sentence, GeneCIS uses single-word instruction to express the user's intent, \textit{e.g.}, focusing/changing a specific object or attribute. From the results, we find that our proposed STiTch achieves the best results in 19/24 cases and outperforms the others at the average score. On the one hand, STiTch preserves rich multimodal knowledge to interpret implicit inputs while filtering out noise through transition. On the other hand, it views the caption as a discrete distribution, showing great potential in capturing diverse visual semantics, leading to fine-grained retrieval for this complex task. 

\subsection{Further Analysis}
In addition to the numerical comparisons on four CIR benchmarks. In this section, we provide further analysis of the ablation results and visualizations of the proposed modules.

\vspace{2mm}
\noindent \textbf{Main component analysis.} 
To evaluate the impacts of each proposed module in STiTch, we report the ablation results in Tab.~\ref{module_ab_tt}. We find that 1) both transition and transportation show a positive improvement compared to the base model (first row, query only); 2) The transition module achieves higher improvements than the transportation module in most cases. This shows the validity of our proposed transition vector 
in assessing
the difference between the reference and target images. It highlights the text modification and offers useful guidance to steer the original caption toward the target image, resulting in more correct alignments. 3) Through collaboration, our complete model achieves the best performance. This highlights the motivation of our STiTch: the unnecessary visual details introduced in original captions and the simple point-based alignment.

\begin{figure}[h]
\centering
\includegraphics[width=0.80\linewidth]{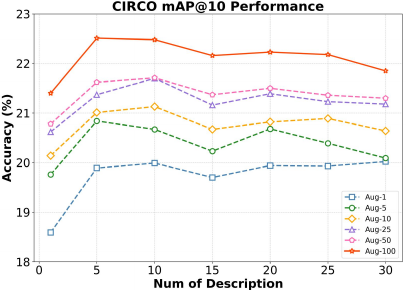}
\caption{\small{Ablation on the number of descriptions and image augmentations (Aug-$10$: $10$ augmentations per image) on the CIRCO dataset with CLIP-B/32.}}
\label{km}
\vspace{-4mm}
\end{figure}

\vspace{2mm}
\noindent \textbf{Impacts of caption number and augmentation views.} 
{We report the study of caption number $K$ and augmentation views $M$ in Eq.~\ref{Q_y} on CIRCO dataset, with results in Fig.~\ref{km}. 
From these results, we first find that the performance shows a large improvement when $K>1$. This demonstrates the effectiveness of using diverse captions, especially after semantic transitions, to capture rich semantic information. Moreover, increasing the number of image augmentations $M$ consistently enhances performance, 
highlighting the value of multi-scale image diversity. We suggest that setting $K = 5$ and $M = 10$ is sufficient to achieve strong performance across most datasets.}

\begin{figure}[h]
\centering
\includegraphics[width=0.80\linewidth]{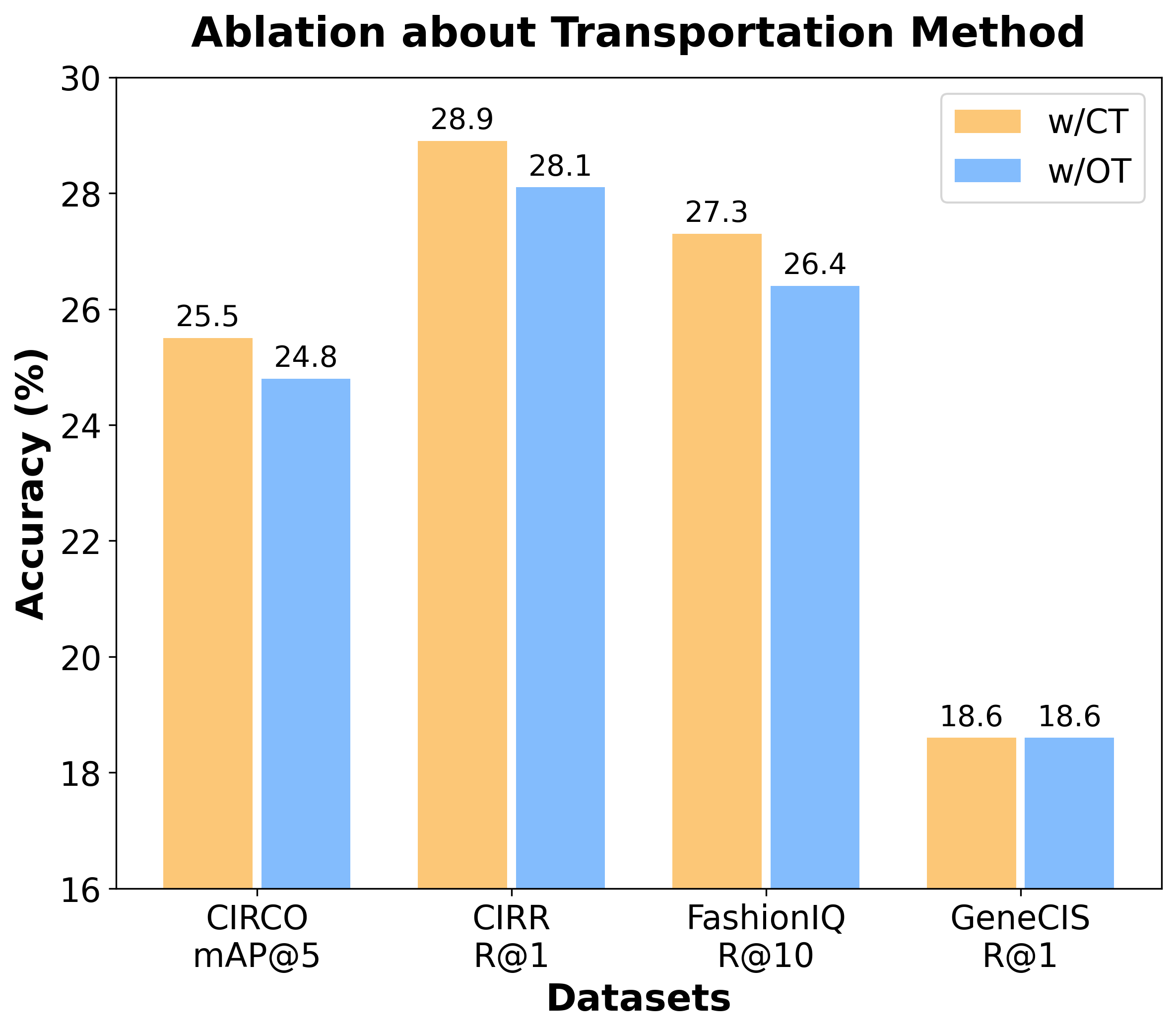}
\vspace{-3mm}
\caption{\small{Ablation results of different alignment strategies across four datasets with CLIP-L/14.}}
\label{ot}
\vspace{-4mm}
\end{figure}

\vspace{2mm}
\noindent \textbf{Impacts of the bidirectional distance.} Recalling that we propose to measure the distance between the modified caption $P_t$ and the target image augmentation $Q_y$ using a bidirectional transportation distance in Eq.~\ref{lbi}. This enables the proposed STiTch 
to not only calculate the transport cost from $P_t$ to $Q_y$ but also consider the reverse cost, showing better alignments across the vision-language domains. {To identify such improvements, we replace the 
bidirectional CT distance with the optimal transport (OT) distance and report the comparison on four datasets in Fig.~\ref{ot}.} Note that the CT distance consistently beats the OT distance in all cases, showing the effectiveness of our bidirectional alignment.

\begin{table*}[!t]
\centering
\caption{\small{Ablation results on the transition and transportation modules. All results are conducted on CIRCO and CIRR datasets with CLIP-bigG/14.}}
\label{module_ab_tt}
\resizebox{0.8\textwidth}{!}{%
\begin{tabular}{cc|cccc||ccc|ccc}
\toprule
\multicolumn{2}{c}{\textbf{CIRCO + CIRR $\rightarrow$}} & \multicolumn{4}{c}{\textbf{CIRCO}} & \multicolumn{6}{c}{\textbf{CIRR}} \\
\hline
\multicolumn{2}{c}{Strategy} & \multicolumn{4}{c}{mAP@k} & \multicolumn{3}{c}{Recall@k} & \multicolumn{3}{c}{$\text{Recall}_{\text{Subset}}$@k} \\
\textit{Transition} & \textit{Transportation} & k=5 & k=10 & k=25 & k=50 & k=1 & k=5 & k=10 & k=1 & k=2 & k=3 \\
\hline
\textcolor{red}{\ding{55}} & \textcolor{red}{\ding{55}} & 31.23 & 32.87 & 36.32 & 38.04 & 37.22 & 67.36 & 77.84 & 69.93 & 86.48 & 94.05\\
\textcolor{red}{\ding{51}} & \textcolor{red}{\ding{55}} & 31.89 & 34.46 & 37.94 & 39.67 & {38.33} & {68.45} & {78.03} & {72.81} & {88.13} & {94.51}\\
\textcolor{red}{\ding{55}} & \textcolor{red}{\ding{51}} & 32.14 & 34.78 & 37.87 & 39.48 & 38.48 & 68.38 & 78.34 & 72.15 & 88.04 & 94.48 \\
\textcolor{red}{\ding{51}} & \textcolor{red}{\ding{51}} & \textbf{34.40} & \textbf{35.56} & \textbf{38.07} & \textbf{40.02} & \textbf{39.23} & \textbf{69.95} & \textbf{79.56} & \textbf{73.56} & \textbf{89.50} & \textbf{95.86}\\
\bottomrule
\label{ablation_ta}
\end{tabular}}
\vspace{-2mm}
\end{table*}

\begin{figure*}[!t]
\centering
\includegraphics[width=0.97\textwidth]{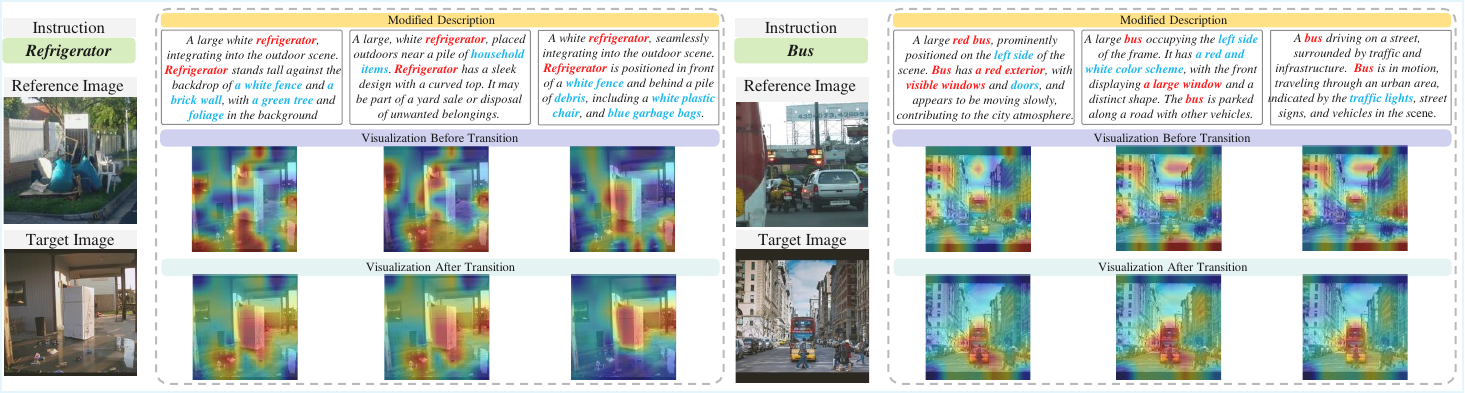} 
\caption{{\small{
Visualization of the GeneCIS dataset on the 'Focus Object' task. Heatmaps before and after the transition on target image are shown. Captions generated by MLLMs often contain irrelevant visual noise (\textit{\sky{blue}} text), while the STiTch model effectively suppresses such noise and highlights the correct focus object (\textit{\rr{red}} text).
}}
\label{vis}}
\end{figure*}

\begin{table*}[!t]
\setlength{\tabcolsep}{8pt} 
\renewcommand{\arraystretch}{0.7} 
\centering
\caption{\small{Efficiency comparison with SOTA methods. Inference times (seconds per query) are reported for each model using a single NVIDIA A6000 GPU.}}
\label{efficiency_vs}
\scalebox{0.9}{
\begin{tabular}{lccccccc}
\toprule
\textbf{Model} & Pic2Word & SEARLE & CIReVL & OSrCIR & LDRE &SEIZE & STiTch \\
\midrule
\textbf{Time (s)} & $<$0.01 & $<$0.01 & 2.97+0.85=3.82 & 6.65 & 1.30+2.98=4.28 &5.8+4.5=10.3 & 3.5 \\
\bottomrule
\end{tabular}}
\vspace{-2mm}
\end{table*}

\vspace{1.5mm}
\noindent \textbf{Visualization Analysis.} From the above analysis, we find that the transition module plays a key role in improving the CIR performance. It estimates the relative vector as the difference between the reference and target image and provides correct information to refine the generated captions. To make a clearer understanding, we visualize the heat maps of two samples from the GeneCIS dataset on the ``Focus Object" task in Fig.~\ref{vis}. Here, we directly calculate the cosine similarity between the visual patch embeddings and the caption embedding as the score of the heat maps. The first row denotes three generated captions and the last two heat maps denote the corresponding visualization results before and after the transition module. We find that the generated captions indeed introduce visual noise for the CIR task. For example, the retrieval attentions are often disturbed by visual noise such as ``white fence", ``white color" and ``left side of the scene". In contrast, the visualizations of STiTch often focus on the correct object, leading to higher CIR performance.

\vspace{1.5mm}
\noindent \textbf{Efficiency Analysis.} Tab. \ref{efficiency_vs} shows the inference times (seconds per query) of various methods. Overall, LLM-based methods generally incur higher test-time latency than traditional mapping models due to the query stage. For instance, CIReVL, LDRE and SEIZE are two-stage approaches that require significant time to generate target descriptions, while OSrCIR involves extensive chain-of-thought reasoning during inference. In contrast, our STiTch achieves the lowest test time among LLM-based methods when taking image augmentation into consideration, showing the efficiency of our collaborative modules.

\section{Conclusion}
\label{sec:conclusion}

In this paper, we propose STiTch, a novel one-stage, training-free framework for the zero-shot composed image retrieval task. STiTch improves the quality of the generated caption by MLLMs via a transition vector and views the captions and target image as two discrete distributions in the embedding space. A bidirectional transport distance is developed to measure the similarity across the vision-language domains. Our approach not only achieves strong performance on four CIR benchmarks but also, provides interpretability via the visualization of the transferred caption and the target images. Extensive ablations also demonstrate the effectiveness of STiTch. Since its natural flexibility and simplicity, we hope STiTch provides innovative ideas for follow-up studies.

\clearpage
\subsection*{Acknowledgments}
This research was supported in part by the Hong Kong RGC General Research Fund (Grant Nos. 15221123, 15216424, and 15211525) and the PolyU Internal Research Fund (Grant Nos. P0058468 and P0056171); in part by the Young Scientists Fund of the National Natural Science Foundation of China (Grant No. 62506237); in part by the National Natural Science Foundation of China (Grant No. 62576215); and in part by the Scientific Foundation for Youth Scholars of Shenzhen University.
\normalem
{
    \small
    \bibliographystyle{ieeenat_fullname}
    \bibliography{main}
}

\clearpage
\setcounter{page}{1}
\maketitlesupplementary
\renewcommand{\thesubsection}{\Alph{subsection}}

\newenvironment{tightitem}{
  \begin{itemize}
    \setlength{\itemsep}{2pt}
    \setlength{\parskip}{0pt}
    \setlength{\parsep}{0pt}
}{\end{itemize}}

\newenvironment{tightsubitem}{
  \begin{itemize}
    \setlength{\itemsep}{1pt}
    \setlength{\parskip}{0pt}
    \setlength{\parsep}{0pt}
}{\end{itemize}}

\newcounter{apxctr}
\setcounter{apxctr}{0}

\section*{\underline{Appendix Overview}}

The supplementary material is organized into the following sections:
\begin{tightitem}

    \stepcounter{apxctr}
    \item \textbf{\hyperref[algorithm]{\Alph{apxctr}. Algorithm of STiTch Process}}
    
    \stepcounter{apxctr}
    \item \textbf{\hyperref[add_res]{\Alph{apxctr}. Additional Comparative Results}}

    \stepcounter{apxctr}
    \item \textbf{\hyperref[appendix_sec_mllm]{\Alph{apxctr}. Impacts of Different MLLMs}}

    \stepcounter{apxctr}
    \item \textbf{\hyperref[add_ab]{\color{cvprblue}\Alph{apxctr}. Additional Ablation Experiments}}

        \begin{tightsubitem}
            \item \hyperref[bid]{Impacts of the Bidirectional Distance}
            \item \hyperref[cap_aug]{Impacts of Caption Number and Augmentation Views}
            \item \hyperref[hyper]{Hyper-parameters Study}
        \end{tightsubitem}

    \stepcounter{apxctr}
    \item \textbf{\hyperref[vis]{\Alph{apxctr}. More Visualization}}

    \stepcounter{apxctr}
    \item \textbf{\hyperref[seize]{\Alph{apxctr}. Further Comparison with SEIZE}}

    \stepcounter{apxctr}
    \item \textbf{\hyperref[in_context]{\Alph{apxctr}. STiTch In-Context Learning Details}}
    
    \stepcounter{apxctr}
    \item \textbf{\hyperref[future]{\Alph{apxctr}. Limitations and Future Work}}


\end{tightitem}


\subsection{Algorithm of STiTch Process}
\label{algorithm}

\begin{algorithm}[!ht]
\footnotesize
\caption{\small Inference algorithm of STiTch.}
\label{alg}

\begin{algorithmic}[1]

\STATE \textbf{Input:} reference image $x$, text modification $m$, 
target image database $\mathbf{Y}=\{y_n\}_{n=1}^N$, 
a pre-trained CLIP model, and a pre-trained MLLM; 
the number of query times $K$, and the number of image augmentations $M$.

\STATE \textbf{Output:} retrieval score $p(y|x,m)$ over all target images.

\STATE \textbf{Querying:} Complete the input prompts with $x$ and $m$, 
and query the MLLM $K$ times to collect descriptions $P_t$ from Eq.~\ref{P_t}.

\STATE \textbf{Transition:} Calculate $\Delta \mathbf{v}$ in Eq.~\ref{Delta} 
by feeding $m$ into the CLIP text encoder, and obtain transferred $P_t$ 
from Eq.~\ref{P_tv2}.

\FOR{image $y_n$ in $\mathbf{Y}$}

    \STATE Collect $Q_y$ in Eq.~\ref{Q_y} by augmenting target image $y_n$ 
    for $M-1$ times.

    \STATE Calculate $\mathcal{L}_{P_t, Q_{y_n}}$ from Eq.~\ref{lbi}.

\ENDFOR

\STATE \textbf{Return:} Calculate retrieval score from Eq.~\ref{qta}.

\end{algorithmic}
\end{algorithm}

\begin{table*}[ht]
\centering
\caption{Performance comparison on CIRCO and CIRR datasets. {{Both ViT-B and ViT-L are loaded from OpenAI official weights, while ViT-G is loaded from OpenCLIP}.}}
\label{appendix_c}
\resizebox{0.9\textwidth}{!}{%
\begin{tabular}{c|c|c|cccc||ccc|ccc}
\toprule
\multicolumn{3}{c}{\textbf{CIRCO + CIRR $\rightarrow$}} & \multicolumn{4}{c}{\textbf{CIRCO}} & \multicolumn{6}{c}{\textbf{CIRR}} \\
\hline
\multicolumn{3}{c}{Metric} & \multicolumn{4}{c}{mAP@k} & \multicolumn{3}{c}{Recall@k} & \multicolumn{3}{c}{$\text{Recall}_{\text{Subset}}$@k} \\
Arch & Method & Train & k=5 & k=10 & k=25 & k=50 & k=1 & k=5 & k=10 & k=1 & k=2 & k=3 \\
\hline
\multirow{5}{*}{ViT-B/32} 
& PALAVRA & \textcolor{red}{\ding{51}} & 4.61 & 5.32 & 6.33 & 6.80 & 16.62 & 43.49 & 58.51 & 41.61 & 65.30 & 80.94 \\
& SEARLE & \textcolor{red}{\ding{51}} & 9.35 & 9.94 & 11.13 & 11.84 & 24.00 & 53.42 & 66.82 & 54.89 & 76.60 & 88.19 \\
& CIReVL & \textcolor{green}{\ding{55}} & 14.94 & 15.42 & 17.00 & 17.82 & 23.94 & 52.51 & 66.00 & 60.17 & 80.05 & 90.19 \\
& LDRE & \textcolor{green}{\ding{55}} & 17.96 & 18.32 & 20.21 & 21.11 & 25.69 & 55.13 & 69.04 & 60.53 & 80.65 & 90.70 \\
& OSrCIR & \textcolor{green}{\ding{55}} & 18.04 & 19.17 & 20.94 & 21.85 & 25.42 & 54.54 & 68.19 & 62.31 & 80.86 & 91.13 \\
& SEIZE & \textcolor{green}{\ding{55}} & \uline{19.04} & \uline{19.64} & \uline{21.55} & \uline{22.49} & \textbf{27.47} & \textbf{57.42} & {70.17} & \uline{65.59} & \textbf{84.48} & \uline{92.77} \\
& \cellcolor{gray!20}\textbf{STiTch(Ours)} & \cellcolor{gray!20}\textcolor{green}{\ding{55}} & \cellcolor{gray!20}\textbf{20.26} & \cellcolor{gray!20}\textbf{21.01} & \cellcolor{gray!20}\textbf{23.01} & \cellcolor{gray!20}\textbf{24.04} & \cellcolor{gray!20}{25.83} & \cellcolor{gray!20}{55.25} & \cellcolor{gray!20}\textbf{70.20 }& \cellcolor{gray!20}\textbf{65.64} & \cellcolor{gray!20}{83.60} &\cellcolor{gray!20}\textbf{92.80} \\
\midrule
\multirow{7}{*}{ViT-L/14} 
& Pic2Word& \textcolor{red}{\ding{51}} & 8.72 & 9.51 & 10.64 & 11.29 & 23.90 & 51.70 & 65.30 & 53.76 & 74.46 & 87.08 \\
& SEARLE & \textcolor{red}{\ding{51}} & 11.68 & 12.73 & 14.33 & 15.12 & 24.24 & 52.48 & 66.29 & 53.76 & 75.01 & 88.19 \\
& LinCIR& \textcolor{red}{\ding{51}} & 12.59 & 13.58 & 15.00 & 15.85 & 25.04 & 53.25 & 66.68 & 57.11 & 77.37 & 88.89 \\
& Context-I2W & \textcolor{red}{\ding{51}} & 13.04 & 14.62 & 16.14 & 17.16 & 25.60 & 55.10 & 68.50 & - & - & - \\
& CIReVL & \textcolor{green}{\ding{55}} & 18.57 & 19.01 & 20.89 & 21.80 & 24.55 & 52.31 & 64.92 & 59.54 & 79.88 & 89.69    \\
& LDRE & \textcolor{green}{\ding{55}} & 23.35 & 24.03 & 26.44 & 27.50 & 26.53 & 55.57 & 67.54 & 60.43 & 80.31 & 89.90 \\
& OSrCIR & \textcolor{green}{\ding{55}} & 23.87 & 25.33 & 27.84 & 28.97 & \textbf{29.45} & 57.68 & {69.86} & 62.12 & 81.92 & 91.10 \\
& SEIZE & \textcolor{green}{\ding{55}} & 24.98 & 25.82 & 28.24 & 28.35 & {28.65} & 57.16 & {69.23} & 62.22 & 84.05 & 92.34 \\
& \cellcolor{gray!20} \textbf{STiTch(Ours)} & \cellcolor{gray!20}\textcolor{green}{\ding{55}} &  \cellcolor{gray!20}\textbf{25.55} & \cellcolor{gray!20}\textbf{26.27} & \cellcolor{gray!20}\textbf{28.81} & \cellcolor{gray!20}\textbf{29.99} & \cellcolor{gray!20}{28.87} & \cellcolor{gray!20}\textbf{57.97} & \cellcolor{gray!20}\textbf{69.90} & \cellcolor{gray!20}\textbf{65.22} & \cellcolor{gray!20}\textbf{84.10} & \cellcolor{gray!20}\textbf{92.37}\\
\midrule
\multirow{6}{*}{ViT-G/14}
&
Pic2Word & \textcolor{red}{\ding{51}} & 5.54 & 5.59 & 6.68 & 7.12 & 30.41 & 58.12 & 69.23 & 68.92 & 85.45 & 93.04 \\
& 
SEARLE & \textcolor{red}{\ding{51}} & 13.20 & 13.85 & 15.32 & 16.04 & 34.80 & 64.07 & 75.11 & 68.72 & 84.70 & 93.23 \\
& 
LinCIR & \textcolor{red}{\ding{51}} & 19.71 & 21.01 & 23.13 & 24.18 & 35.25 & 64.72 & 76.05 & 63.35 & 82.22 & 91.98 \\
& CIReVL & \textcolor{green}{\ding{55}} & 26.77 & 27.59 & 29.96 & 31.03 & 34.65 & 64.29 & 75.06 & 67.95 & 84.87 & 93.21 \\
& LDRE & \textcolor{green}{\ding{55}} & 31.12 & 32.24 & 34.95 & 36.03 & 36.15 & 66.39 & 77.25 & 68.82 & 85.66 & 93.76 \\
& OSrCIR & \textcolor{green}{\ding{55}} & 30.47 & 31.14 & 35.03 & 36.59 & 37.26 & 67.25 & 77.33 & 69.22 & 85.28 & 93.55 \\
& SEIZE & \textcolor{green}{\ding{55}} & 32.46 & 33.77 & 36.46 & 37.55 & 38.87 & 69.42 & 79.42 & \textbf{74.15} & 89.23 & 95.71 \\
& \cellcolor{gray!20} \textbf{STiTch (Ours)} & \cellcolor{gray!20}\textcolor{green}{\ding{55}} & \cellcolor{gray!20}\textbf{34.40} & \cellcolor{gray!20}\textbf{35.56} & \cellcolor{gray!20}\textbf{38.07} & \cellcolor{gray!20}\textbf{40.02} & \cellcolor{gray!20}\textbf{39.23} & \cellcolor{gray!20}\textbf{69.95} & \cellcolor{gray!20}\textbf{79.56} & \cellcolor{gray!20}{73.56} & \cellcolor{gray!20}\textbf{89.50} & \cellcolor{gray!20}\textbf{95.86} \\
\bottomrule
\end{tabular}}
\end{table*}

\begin{table*}[ht]
\centering
\caption{Performance comparison on Fashion-IQ datasets. Both ViT-B and ViT-L are loaded from OpenAI official weights, while ViT-G is loaded from OpenCLIP. (*) denotes we rerun the experiments on the OpenAI weights, and (\dag) denotes we rerun the experiments on the OpenCLIP weights.}
\label{appendix_f}
\resizebox{0.8\textwidth}{!}{%
\begin{tabular}{c|c|c|cccccc|cc}
\toprule
\multicolumn{3}{c}{\textbf{Fashion-IQ $\rightarrow$}} & \multicolumn{2}{c}{\textbf{Shirt}} & \multicolumn{2}{c}{\textbf{Dress}} & \multicolumn{2}{c}{\textbf{Toptee}} & \multicolumn{2}{c}{\textbf{Average}} \\
\hline
Arch & Method & Train & R@10 & R@50 & R@10 & R@50 & R@10 & R@50 & R@10 & R@50 \\
\hline 
\multirow{4}{*}{ViT-B/32} 
& PALAVRA & \textcolor{red}{\ding{51}} & 21.49 & 37.05 & 17.25 & 35.94 & 20.55 & 38.76 & 19.76 & 37.25 \\
& SEARLE & \textcolor{red}{\ding{51}} & 24.44 & 41.61 & 18.54 & 39.51 & 25.70 & 46.46 & 22.89 & 42.53 \\
& CIReVL & \textcolor{green}{\ding{55}}& 28.36 & 47.84 &  25.29 & 46.36 &  31.21 & 53.85 & 28.28 & 49.35 \\
& CIReVL*& \textcolor{green}{\ding{55}} & 22.03 & 37.00 & 13.34& 30.14& 18.97 & 38.19 &  18.11 & 35.11  \\
& CIReVL\dag & \textcolor{green}{\ding{55}} & 27.72 & 46.12 & 22.01& 41.60& 30.09 & 52.22 & 26.60 & 46.64\\
& OSrCIR & \textcolor{green}{\ding{55}} & 31.16 & 51.13 & 29.35 & 50.37 & 36.51 & 58.71 & 32.34 & 53.40 \\
& OSrCIR* & \textcolor{green}{\ding{55}} & 22.77 &  40.87 & 17.01 & 37.04 & 20.75 & 41.00 & 20.18 & 39.60  \\
& OSrCIR\dag & \textcolor{green}{\ding{55}} &32.83  & 52.06 & 29.75 & 51.91 & 36.31 & 58.24 & 32.96 & 54.07 \\
&SEIZE & \textcolor{green}{\ding{55}} & 29.38 & 47.97 & 25.37 & 46.84 & 32.07 & 54.78 & 28.94 & 49.35 \\
&\cellcolor{gray!20}\textbf{STiTch(Ours)} & \cellcolor{gray!20}\textcolor{green}{\ding{55}} & \cellcolor{gray!20}25.22 & \cellcolor{gray!20}44.16 & \cellcolor{gray!20}18.59 & \cellcolor{gray!20}40.16 & \cellcolor{gray!20}25.97 & \cellcolor{gray!20}47.61 & \cellcolor{gray!20}23.26 & \cellcolor{gray!20}43.98 \\
\midrule
\multirow{4}{*}{ViT-G/14} 
& Pic2Word & \textcolor{red}{\ding{51}} & 33.17 & 50.39 & 25.43 & 47.65 & 35.24 & 57.62 & 31.28 & 51.89 \\
& SEARLE & \textcolor{red}{\ding{51}} & 36.46 & 55.35 & 28.16 & 50.32 & 39.83 & 61.45 & 34.81 & 55.71 \\
& LinCIR & \textcolor{red}{\ding{51}} & 46.76 & 65.11 & 38.08 & 60.88 & 50.48 & 71.09 & 45.11 & 65.69 \\
& CIReVL & \textcolor{green}{\ding{55}} &  29.85 & 51.07 & 27.07 & 49.53 & 35.80 & 56.14 & 32.19 & 52.36 \\
& CIReVL* & \textcolor{green}{\ding{55}} & 31.65 & 49.07 &  23.90 & 43.13 & 32.53 & 53.19 & 29.36 & 48.46\\
& CIReVL\dag & \textcolor{green}{\ding{55}} & 32.63 & 50.05 & 25.09 & 45.12 & 34.42 & 55.12 & 30.71 & 50.10 \\
& OSrCIR & \textcolor{green}{\ding{55}} & 38.65 & 54.71 & 33.02 & 54.78 & 41.04 & 61.83 & 37.57 & 57.11 \\
& OSrCIR* & \textcolor{green}{\ding{55}}  & 36.56 & 55.45 & 30.69 & 53.25 & 40.13 & 61.30 & 35.79 & 56.67\\
& OSrCIR\dag & \textcolor{green}{\ding{55}}  & 37.39 & 56.92 & 30.59 & 53.50 &  39.72 & 61.04 & 35.79 & 57.15 \\
& SEIZE & \textcolor{green}{\ding{55}} & 43.60 & 65.42 & 39.61 & 61.02 & 45.94 & 71.12 & 43.05 & 65.85 \\
& \cellcolor{gray!20}\textbf{STiTch(Ours)} & \cellcolor{gray!20}\textcolor{green}{\ding{55}} & \cellcolor{gray!20}{39.48}  & \cellcolor{gray!20}{56.59} & \cellcolor{gray!20}{35.04} & \cellcolor{gray!20}{56.74} & \cellcolor{gray!20}{42.86} & \cellcolor{gray!20}{64.95} & \cellcolor{gray!20}{39.12} &  \cellcolor{gray!20}{59.43} \\
\bottomrule
\end{tabular}}
\end{table*}

\begin{table*}[ht]
\centering
\caption{Performance comparison on GeneCIS datasets. \textbf{Both ViT-B and ViT-L are loaded from openai official weights, while ViT-G is loaded from openclip}.}
\label{appendix_g}
\resizebox{\textwidth}{!}{%
\begin{tabular}{c|c|c|cccccccccccc|c}
\toprule
\multicolumn{3}{c}{\textbf{GeneCIS $\rightarrow$}} & \multicolumn{3}{c}{\textbf{Focus Attribute}} & \multicolumn{3}{c}{\textbf{Change Attribute}} & \multicolumn{3}{c}{\textbf{Focus Object}} & \multicolumn{3}{c}{\textbf{Change Object}} & \multicolumn{1}{c}{\textbf{Average}}  \\
\hline
Arch & Method & Train & R@1 & R@2 & R@3 & R@1 & R@2 & R@3 & R@1 & R@2 & R@3 & R@1 & R@2 & R@3 & R@1 \\
\hline
\multirow{5}{*}{ViT-B/32} 
& SEARLE & \textcolor{red}{\ding{51}} & 18.9 & 30.6 & 41.2 & 13.0 & 23.8 & 33.7 & 12.2 & 23.0 & 33.3 & 13.6 & 23.8 & 33.3 & 14.4 \\
& CIReVL & \textcolor{green}{\ding{55}} & 17.9 & 29.4 & 40.4 & 14.8 & 25.8 & 35.8 & 14.6 & 24.3 & 33.3 & 16.1 & {27.8} & 37.6 & 15.9 \\
& OSrCIR & \textcolor{green}{\ding{55}} & {19.4} & {32.7} & {42.8} & {16.4} & {27.7} & {38.1} & {15.7} & {25.7} & {35.8} & {18.2} & {30.1} & {39.4} & {17.4} \\
& \cellcolor{gray!20}\textbf{STiTch(Ours)} & \cellcolor{gray!20}\textcolor{green}{\ding{55}} & \cellcolor{gray!20}\textbf{21.1} & \cellcolor{gray!20}\textbf{35.0} & \cellcolor{gray!20}\textbf{45.5} & \cellcolor{gray!20}\textbf{17.9} & \cellcolor{gray!20}\textbf{29.9} & \cellcolor{gray!20}\textbf{40.4} & \cellcolor{gray!20}\textbf{16.4} & \cellcolor{gray!20}\textbf{28.5} & \cellcolor{gray!20}\textbf{38.9} & \cellcolor{gray!20}\textbf{18.3} & \cellcolor{gray!20}\textbf{30.1} & \cellcolor{gray!20}\textbf{39.5} & \cellcolor{gray!20}\textbf{18.4} \\
\midrule
\multirow{7}{*}{ViT-L/14} 
& SEARLE& \textcolor{red}{\ding{51}} & 17.1 & 29.6 & 40.7 & 16.3 & 25.2 & 34.2 & 12.0 & 22.2 & 30.9 & 12.0 & 24.1 & 33.9 & 14.4 \\
& LinCIR & \textcolor{red}{\ding{51}} & 16.9 & 30.0 & 41.5 & 16.2 & 28.0 & 36.8 & 8.3 & 17.4 & 26.2 & 7.4 & 15.7 & 25.0 & 12.2 \\
& Context-I2W & \textcolor{red}{\ding{51}} & 17.2 & 30.5 & 41.7 & 16.4 & 28.3 & 37.1 & 8.7 & 17.9 & 26.9 & 7.7 & 16.0 & 25.4 & 12.7\\
& CIReVL & \textcolor{green}{\ding{55}} & 19.5 & 31.8 & 42.0 & 14.4 & 26.0 & 35.2 & 12.3 & 21.8 & 30.5 & 17.2 & 28.9 & 37.6 & 15.9\\
& OSrCIR & \textcolor{green}{\ding{55}} & \textbf{20.9} & \uline{33.1} & \uline{44.5} & \uline{17.2} & \uline{28.5} & \uline{37.9} & \uline{15.0} & \uline{23.6} & \uline{34.2} & \uline{18.4} & \uline{30.6} & \uline{38.3} & \uline{17.9} \\
& SEIZE & \textcolor{green}{\ding{55}} & {20.5} & {33.4} & {45.0} & {17.6} & {28.9} & {38.5} & {15.4} & {25.6} & {36.2} & {18.7} & {30.9} & {39.8} & {18.1} \\
& \cellcolor{gray!20}\textbf{STiTch(Ours)} & \cellcolor{gray!20}\textcolor{green}{\ding{55}} & \cellcolor{gray!20}\uline{20.3} & \cellcolor{gray!20}\textbf{34.6} & \cellcolor{gray!20}\textbf{46.4} & \cellcolor{gray!20}\textbf{18.3} & \cellcolor{gray!20}\textbf{29.8} & \cellcolor{gray!20}\textbf{41.6} & \cellcolor{gray!20}\textbf{16.8} & \cellcolor{gray!20}\textbf{28.5} & \cellcolor{gray!20}\textbf{38.4} & \cellcolor{gray!20}\textbf{18.8} & \cellcolor{gray!20}\textbf{31.0} & \cellcolor{gray!20}\textbf{40.3} & \cellcolor{gray!20}\textbf{18.6} \\
\midrule
\multirow{5}{*}{ViT-G/14} 
& LinCIR & \textcolor{red}{\ding{51}} & 19.1 & 33.0 & 42.3 & 17.6 & 30.2 & 38.1 & 10.1 & 19.1 & 28.1 & 7.9 & 16.3 & 25.7 & 13.7\\
& CIReVL & \textcolor{green}{\ding{55}} & 20.5 & 34.0 & 44.5 & 16.1 & 28.6 & 39.4 & 14.7 & 25.2 & 33.0 & 18.1 & 31.2 & 41.0 & 17.4\\
& OSrCIR & \textcolor{green}{\ding{55}} & {22.7} & {36.4} & \uline{47.0} & \uline{17.9} & \uline{30.8} & \uline{42.0} & \uline{16.9} & \uline{28.4} & \uline{36.7} & \textbf{21.0} & \textbf{33.4} & \textbf{44.2} & \uline{19.6} \\
& SEIZE & \textcolor{green}{\ding{55}} & \textbf{22.9} & 36.2 & {47.3} & {18.6} & {31.4} & {42.7} & {18.2} & {28.8} & {37.6} & {19.6} & {33.0} & {43.5} & {19.8} \\
& \cellcolor{gray!20}\textbf{STiTch(Ours)} & \cellcolor{gray!20}\textcolor{green}{\ding{55}} & \cellcolor{gray!20}\uline{21.9} & \cellcolor{gray!20}\textbf{36.4} & \cellcolor{gray!20}\textbf{47.9} & \cellcolor{gray!20}\textbf{19.6} & \cellcolor{gray!20}\textbf{31.9} & \cellcolor{gray!20}\textbf{42.8} & \cellcolor{gray!20}\textbf{20.2} & \cellcolor{gray!20}\textbf{30.3} & \cellcolor{gray!20}\textbf{39.6} & \cellcolor{gray!20}{19.7} & \cellcolor{gray!20}{33.2} & \cellcolor{gray!20}\uline{43.4} & \cellcolor{gray!20}\textbf{20.4} \\
\bottomrule
\end{tabular}}
\end{table*}

\begin{table*}[ht]
\centering
\caption{Ablation results of different alignment strategies on CIRCO and CIRR datasets.}
\label{appendix_ot}
\resizebox{0.85\textwidth}{!}{%
\begin{tabular}{c|c|cccc||ccc|ccc}
\toprule
\multicolumn{2}{c}{\textbf{CIRCO + CIRR $\rightarrow$}} & \multicolumn{4}{c}{\textbf{CIRCO}} & \multicolumn{6}{c}{\textbf{CIRR}} \\
\hline
\multicolumn{2}{c}{Metric} & \multicolumn{4}{c}{mAP@k} & \multicolumn{3}{c}{Recall@k} & \multicolumn{3}{c}{$\text{Recall}_{\text{Subset}}$@k} \\
Arch & Method & k=5 & k=10 & k=25 & k=50 & k=1 & k=5 & k=10 & k=1 & k=2 & k=3 \\
\hline
\multirow{2}{*}{ViT-B/32 (OpenAI)} 
& \cellcolor{gray!20}w/CT & \cellcolor{gray!20}20.26 & \cellcolor{gray!20}21.01 & \cellcolor{gray!20}23.01 & \cellcolor{gray!20}24.04 & \cellcolor{gray!20}25.83 & \cellcolor{gray!20}55.18 & \cellcolor{gray!20}68.22 & \cellcolor{gray!20}65.64 & \cellcolor{gray!20}83.60 & \cellcolor{gray!20}92.80 \\
& w/OT & 19.81 & 20.48 & 22.33 & 23.31 & 24.72 & 53.66 & 66.77 & 65.23 & 83.64 & 92.46 \\
\midrule
\multirow{2}{*}{ViT-B-32 (OpenCLIP)} 
& \cellcolor{gray!20}w/CT& \cellcolor{gray!20}28.21 & \cellcolor{gray!20}28.99 & \cellcolor{gray!20}31.31 & \cellcolor{gray!20}32.53 & \cellcolor{gray!20}32.56 & \cellcolor{gray!20}62.10 & \cellcolor{gray!20}73.86 & \cellcolor{gray!20}70.00 & \cellcolor{gray!20}86.60 & \cellcolor{gray!20}94.87 \\
& w/OT & 23.94 & 24.79 & 27.01 & 28.09 & 31.37 & 60.97 & 72.92 & 69.54 & 85.76 & 94.27 \\
\midrule
\multirow{2}{*}{ViT-L/14 (OpenAI)} 
& \cellcolor{gray!20}w/CT & \cellcolor{gray!20}25.55 & \cellcolor{gray!20}26.27 & \cellcolor{gray!20}28.81 & \cellcolor{gray!20}29.99 & \cellcolor{gray!20}28.87 & \cellcolor{gray!20}57.97 & \cellcolor{gray!20}69.90  & \cellcolor{gray!20}65.22 & \cellcolor{gray!20}84.10 & \cellcolor{gray!20}92.37 \\
& w/OT & 24.76 & 25.85 & 28.52 & 29.67 & 28.05 & 57.01 & 69.64 & 65.71 & 83.74 & 92.17 \\
\midrule
\multirow{2}{*}{ViT-L-14 (OpenCLIP)} 
& \cellcolor{gray!20}w/CT& \cellcolor{gray!20}32.31 & \cellcolor{gray!20}33.33 & \cellcolor{gray!20}36.32 & \cellcolor{gray!20}37.49 & \cellcolor{gray!20}35.04 & \cellcolor{gray!20}65.57 & \cellcolor{gray!20}76.41 & \cellcolor{gray!20}71.52 & \cellcolor{gray!20}88.00 & \cellcolor{gray!20}94.65 \\
& w/OT & 30.23 & 31.22 & 34.13 & 35.20 & 34.39 & 64.48 & 76.17 & 71.98 & 88.34 & 94.68 \\
\bottomrule
\end{tabular}}
\end{table*}


\begin{table*}[ht]
\centering
\caption{Ablation results of different alignment strategies on GeneCIS dataset.}
\label{appendix_ot2}
\resizebox{\textwidth}{!}{%
\begin{tabular}{c|c|cccccccccccc|c}
\toprule
\multicolumn{2}{c}{\textbf{GeneCIS $\rightarrow$}} & \multicolumn{3}{c}{\textbf{Focus Attribute}} & \multicolumn{3}{c}{\textbf{Change Attribute}} & \multicolumn{3}{c}{\textbf{Focus Object}} & \multicolumn{3}{c}{\textbf{Change Object}} & \multicolumn{1}{c}{\textbf{Average}}  \\
\hline
Arch & Method & R@1 & R@2 & R@3 & R@1 & R@2 & R@3 & R@1 & R@2 & R@3 & R@1 & R@2 & R@3 & R@1 \\
\hline
\multirow{2}{*}{ViT-B/32 (OpenAI)} 
& \cellcolor{gray!20}w/CT & \cellcolor{gray!20}21.1 & \cellcolor{gray!20}35.0 & \cellcolor{gray!20}45.5 & \cellcolor{gray!20}17.9 & \cellcolor{gray!20}29.9 & \cellcolor{gray!20}40.4 & \cellcolor{gray!20}16.4 & \cellcolor{gray!20}28.5 & \cellcolor{gray!20}38.9 & \cellcolor{gray!20}18.3 & \cellcolor{gray!20}30.1 & \cellcolor{gray!20}39.5 & \cellcolor{gray!20}18.4 \\
& w/OT & 20.5 & 34.1 & 44.9 & 17.5 & 30.2 & 40.1 & 16.3 & 28.0 & 38.3 & 18.3 & 29.9 & 39.4 & 18.2 \\
\midrule
\multirow{2}{*}{ViT-B-32 (OpenCLIP)} 
& \cellcolor{gray!20}w/CT & \cellcolor{gray!20}20.8 & \cellcolor{gray!20}33.6 & \cellcolor{gray!20}44.2 & \cellcolor{gray!20}17.6 & \cellcolor{gray!20}29.4 & \cellcolor{gray!20}39.7 & \cellcolor{gray!20}17.4 & \cellcolor{gray!20}30.7 & \cellcolor{gray!20}40.4 & \cellcolor{gray!20}19.6 & \cellcolor{gray!20}33.6 & \cellcolor{gray!20}44.2 & \cellcolor{gray!20}18.9 \\
& w/OT & 20.2 & 33.4 & 43.8 & 17.1 & 29.4 & 39.0 & 17.0 & 29.7 & 39.9 & 20.0 & 33.4 & 43.8 & 18.6 \\
\midrule
\multirow{2}{*}{ViT-L/14 (OpenAI)} 
& \cellcolor{gray!20}w/CT & \cellcolor{gray!20}20.3 & \cellcolor{gray!20}34.6 & \cellcolor{gray!20}46.4 & \cellcolor{gray!20}18.3 & \cellcolor{gray!20}29.8 & \cellcolor{gray!20}41.6 & \cellcolor{gray!20}16.8 & \cellcolor{gray!20}28.5 & \cellcolor{gray!20}38.4 & \cellcolor{gray!20}18.8 & \cellcolor{gray!20}31.0 & \cellcolor{gray!20}40.3 & \cellcolor{gray!20}18.6 \\
& w/OT & 20.6 & 34.5 & 45.8 & 18.0 & 29.2 & 40.3 & 16.8 & 27.9 & 38.2 & 18.8 & 30.2 & 40.1 & 18.5 \\
\midrule
\multirow{2}{*}{ViT-L-14 (OpenCLIP)} 
& \cellcolor{gray!20}w/CT & \cellcolor{gray!20}20.5 & \cellcolor{gray!20}33.8 & \cellcolor{gray!20}44.2 & \cellcolor{gray!20}18.1 & \cellcolor{gray!20}29.0 & \cellcolor{gray!20}40.2 & \cellcolor{gray!20}18.5 & \cellcolor{gray!20}29.4 & \cellcolor{gray!20}39.1 & \cellcolor{gray!20}19.9 & \cellcolor{gray!20}32.9 & \cellcolor{gray!20}42.8 & \cellcolor{gray!20}19.3 \\
& w/OT & 20.3 & 33.4 & 44.4 & 17.6 & 28.8 & 39.9 & 17.9 & 28.6 & 38.1 & 19.6 & 33.1 & 42.3 & 18.9 \\
\bottomrule
\end{tabular}}
\end{table*}

\subsection{Additional Comparative Results.}
\label{add_res}
We in this section included more comprehensive comparisons with more methods across various architectures on all datasets presented in Tab.\ref{appendix_c}, Tab.\ref{appendix_f}, and Tab.\ref{appendix_g}. It should be noted that in Tab.\ref{appendix_f}, the notation (*) indicates that we reproduced the experiments using the OpenAI weights, and the (\dag) indicates that we reproduced the experiments using the OpenCLIP weights, respectively.  From these comparisons, our approach outperforms all the baselines in most cases, showing the efficiency of STiTch's three operations.

\subsection{Impacts of Different MLLMs} \label{appendix_sec_mllm}
Like previous works that employ MLLMs to analyze multimodal inputs and generate target descriptions, we specify Qwen2-VL-7B as the MLLM in earlier experiments. Here, we further explore the performance of STiTch with different MLLMs. Specifically, we report the results on Qwen2-VL-2B, Qwen2-VL-7B, LLaVA-Next-7B, and GPT-4o(mini) in Tab.~\ref{append_mllm}. The results show that our STiTch can be applied to MLLMs with different architectures and that the performance improves as the number of MLLM's parameters increases. This demonstrates the potential of STiTch in flexibility and scalability, as it serves as a plug-and-play pipeline that can seamlessly integrate with various MLLMs. Indeed, we observe that different MLLMs can lead to variations in the generated captions and thus impact retrieval results. This observation further supports our core motivation: rather than re-training or fine-tuning the large models, we aim to design a framework that maximizes retrieval effectiveness given any off-the-shelf MLLM.

In addition to Tab.\ref{module_ab_tt} that ablates each module on Qwen-7B, we also report the results with another MLLM GPT-4o(mini) in Tab.\ref{moudle_ab_tt_2}. The ablations on two MLLMs can show the real efficiency of STiTch's modules: (1) \textit{Strategic Synergy Over Raw MLLM Power}: The highest mAP@k values (e.g., 38.93 @k=5, 44.46 @k=50) occur when both Transition and Transportation are enabled. This indicates that STiTch's strength lies in its systematic collaboration of strategies rather than relying solely on MLLM capabilities. Even with the same MLLM (e.g., GPT-4o(mini)), disabling either strategy reduces performance (e.g., Transportation only yields 36.61 @k=5; Transition only yields 37.50 @k=5), confirming that STiTch actively improves task-specific reasoning. (2) \textit{Modular Adaptability}: The results implies STiTch’s strategies are architecture-agnostic. While the choice of MLLM impacts absolute performance, the framework’s relative gains from Transition+Transportation collaboration remain consistent.

\begin{table}[htbp]
\centering
\caption{\small{Ablation results on the transition and transportation modules. All results are conducted on CIRCO datasets with GPT-4o(mini).}}
\label{moudle_ab_tt_2}
\resizebox{0.45\textwidth}{!}{%
\begin{tabular}{cc|cccc}
\toprule
\multicolumn{2}{c}{Strategy} & \multicolumn{4}{c}{mAP@k}\\
\hline
\textit{Transition} & \textit{Transportation} & k=5 & k=10 & k=25 & k=50 \\
\hline
\textcolor{red}{\ding{55}} & \textcolor{red}{\ding{55}} & 35.49 & 37.05 & 40.02 & 41.28\\
\textcolor{red}{\ding{51}} & \textcolor{red}{\ding{55}} & 37.50 & 39.10 & 42.24 & 43.50\\
\textcolor{red}{\ding{55}} & \textcolor{red}{\ding{51}} & 36.61 & 38.10 & 41.11 & 42.38 \\
\textcolor{red}{\ding{51}} & \textcolor{red}{\ding{51}} & \textbf{38.93} & \textbf{40.14} & \textbf{43.18} & \textbf{44.46} \\
\bottomrule
\label{ablation_ta}
\end{tabular}}
\end{table}

\begin{table*}[ht]
\centering
\caption{\small{Performance comparison on CIRCO and CIRR datasets with various MLLMs.}}
\label{append_mllm}
\resizebox{0.8\textwidth}{!}{%
\begin{tabular}{c|cccc||ccc|ccc}
\toprule
\multicolumn{1}{c}{\textbf{CIRCO + CIRR $\rightarrow$}} & \multicolumn{4}{c}{\textbf{CIRCO}} & \multicolumn{6}{c}{\textbf{CIRR}} \\
\hline
\multicolumn{1}{c}{Metric} & \multicolumn{4}{c}{mAP@k} & \multicolumn{3}{c}{Recall@k} & \multicolumn{3}{c}{$\text{Recall}_{\text{Subset}}$@k} \\
 Method & k=5 & k=10 & k=25 & k=50 & k=1 & k=5 & k=10 & k=1 & k=2 & k=3 \\
\hline
Qwen-2B & 22.49 & 23.64 & 25.90 & 26.95 & 26.05 & 53.28 & 65.59 & 64.53 & 82.46 & 91.25 \\ 
\cellcolor{gray!20}Qwen-7B & \cellcolor{gray!20}25.55 & \cellcolor{gray!20}26.27 & \cellcolor{gray!20}28.81 & \cellcolor{gray!20}29.99 & \cellcolor{gray!20}28.87 & \cellcolor{gray!20}57.97 & \cellcolor{gray!20}69.90  & \cellcolor{gray!20}65.22 & \cellcolor{gray!20}84.10 & \cellcolor{gray!20}92.37 \\
LLaVA-Next (Mistral-7B) & 24.17 & 24.73 & 27.03 & 28.11 & 26.97 & 55.10 & 66.92 & 65.01 & 82.75 & 91.40 \\
GPT-4o(mini) & 25.68 & 26.50 & 29.16 & 30.30 & 28.59 & 58.13 & 69.99 & 66.15 & 84.98 & 92.86 \\
\bottomrule
\end{tabular}}
\end{table*}

\begin{table*}[ht]
\centering
\caption{Ablation study on CIRCO and CIRR datasets with different number of image augmentation on CLIP-B/32 and fix the number of description to 5.}
\label{appendix_km}
\resizebox{0.75\textwidth}{!}{%
\begin{tabular}{c|cccc||ccc|ccc}
\toprule
\multicolumn{1}{c}{\textbf{CIRCO + CIRR $\rightarrow$}}  & \multicolumn{4}{c}{\textbf{CIRCO}} & \multicolumn{6}{c}{\textbf{CIRR}} \\
\hline
\multicolumn{1}{c}{Metrics} & \multicolumn{4}{c}{mAP@k} & \multicolumn{3}{c}{Recall@k} & \multicolumn{3}{c}{$\text{Recall}_{\text{Subset}}$@k} \\
 Num & k=5 & k=10 & k=25 & k=50 & k=1 & k=5 & k=10 & k=1 & k=2 & k=3 \\
\hline
 1  & 19.73 & 19.89 & 21.68 & 22.63 & 25.16 & 53.59 & 66.46 & 63.88 & 82.87 & 92.19 \\
 5  & 20.19 & 20.84 & 22.70 & 23.73 & 25.25 & 54.00 & 67.40 & 64.46 & 83.61 & 92.39 \\
 \cellcolor{gray!20}10 & \cellcolor{gray!20}20.26 & \cellcolor{gray!20}21.01 & \cellcolor{gray!20}23.01 & \cellcolor{gray!20}24.04 & \cellcolor{gray!20}25.83 & \cellcolor{gray!20}55.25 & \cellcolor{gray!20}68.22 & \cellcolor{gray!20}65.64 & \cellcolor{gray!20}83.60 & \cellcolor{gray!20}92.80\\
 25 & 20.60 & 21.37 & 23.55 & 24.55 & 25.64 & 55.45 & 68.87 & \uline{65.71} & \uline{84.41} & 92.46 \\
 50 & \uline{21.01} & \uline{21.62} & \uline{23.74} & \uline{24.79} & \textbf{26.15} & \textbf{55.78} & \textbf{69.16} & \textbf{66.17} & \textbf{84.74} & \uline{93.06} \\
 100 & \textbf{21.96} & \textbf{22.51} & \textbf{24.60} & \textbf{25.62} & \uline{25.67} & \uline{55.69} & \uline{69.08} & 65.45 & 84.36 & \textbf{93.08} \\
\bottomrule
\end{tabular}}
\end{table*}

\begin{table*}[!ht]
\centering
\caption{Sensitivity analysis of $\alpha$ on Qwen2-VL-7B and ViT-B/32 on CIRCO and CIRR datasets (default $\alpha=0.45$ in our main manuscript).}
\label{ab_alpha}
\resizebox{0.70\textwidth}{!}{%
\begin{tabular}{c|cccc||ccc|cc}
\toprule
\multicolumn{1}{c}{\textbf{CIRCO + CIRR $\rightarrow$}}  & \multicolumn{4}{c}{\textbf{CIRCO}} & \multicolumn{5}{c}{\textbf{CIRR}} \\
\hline
\multicolumn{1}{c}{Metrics} & \multicolumn{4}{c}{mAP@k} & \multicolumn{3}{c}{Recall@k} & \multicolumn{2}{c}{$\text{Recall}_{\text{Subset}}$@k} \\
$\alpha$ value & k=5 & k=10 & k=25 & k=50 & k=1 & k=5 & k=10 & k=1 & k=2 \\
\hline
0.1  & 18.37 & 19.09 & 20.77 & 21.76 & 23.28 & 49.98 & 62.36 & 64.05 & 83.21 \\
0.2  & 19.74 & 20.49 & 22.34 & 23.32 & 24.63 & 52.46 & 65.45 & 64.89 & 83.40 \\
0.3  & 21.71 & 22.36 & 24.33 & 25.26 & 25.35 & 54.12 & 67.28 & 65.49 & 83.74 \\
0.4  & 20.73 & 21.37 & 23.33 & 24.37 & 25.81 & 55.37 & 68.34 & 65.23 & 83.67 \\
 \cellcolor{gray!20}0.45 &  \cellcolor{gray!20}20.26 &  \cellcolor{gray!20}21.01 &  \cellcolor{gray!20}23.01 &  \cellcolor{gray!20}24.04 &  \cellcolor{gray!20}25.83 &  \cellcolor{gray!20}55.25 &  \cellcolor{gray!20}68.22 &  \cellcolor{gray!20}65.64 &  \cellcolor{gray!20}83.60 \\
0.5  & 21.47 & 22.47 & 24.46 & 25.46 & 26.02 & 55.45 & 68.58 & 64.82 & 83.49 \\
0.6  & 19.77 & 20.45 & 22.55 & 23.48 & 25.62 & 55.40 & 68.22 & 63.64 & 83.13 \\
0.7  & 19.05 & 20.18 & 22.11 & 23.20 & 25.11 & 54.65 & 68.22 & 63.62 & 82.68 \\
\bottomrule
\end{tabular}}
\end{table*}

\subsection{Additional Ablation Experiments}
\subsubsection{Impacts of the bidirectional distance.}
\label{bid}

To conduct a more comprehensive analysis of the impacts of the bidirectional distance, we supplemented experiments with STiTch under different backbones using CT distance and OT distance as alignment strategy in Tab.\ref{appendix_ot}
and Tab.\ref{appendix_ot2}. The results show that CT outperforms OT, highlighting the advantages of bidirectional fine-grained alignment.

\subsubsection{Impacts of caption number and augmentation views.}
\label{cap_aug}

Moreover, for clarity, we have provided the specific values corresponding to Fig.\ref{km} in the main text and supplemented the results of ablation experiments under different architectures, which can be found in Tab.\ref{appendix_km}. It is evident that compared to a single caption ($k$=1), multiple captions can provide richer multi-modal knowledge to better understand the implicit input, leading to more accurate descriptions.

\subsubsection{Hyper-parameters Study}
\label{hyper}
We report a sensitivity analysis of $\alpha$ in Tab.\ref{ab_alpha}. The results show that STiTch exhibits moderate sensitivity to $\alpha$, with performance being non-monotonic. Specifically, values in the range of $0.3$ – $0.5$ yield optimal results, while overly small or large values degrade performance. This confirms the effectiveness of treating modification as a transition vector, as it helps mitigate biases between MLLM-generated captions and images. For practical use, in accuracy-critical tasks (e.g., CIRCO), we suggest $\alpha\leq0.5$ to avoid over-modification; In recall-critical tasks (e.g., CIRR), starting with $\alpha=0.4$ is reasonable. For new datasets, a grid search within $[0.3,0.5]$ could be conducted, selecting the optimal $\alpha$ based on validation performance tailored to the application's specific needs.

\subsection{More Visualization}
\label{vis}
For a more comprehensive qualitative analysis, we present the visualization results of GeneCIS datasets about the task of focus in Fig. \ref{appendix_vis}. It illustrated that the original generated descriptions indeed introduce visual noise while our STiTch often focuses on the correct object, leading to
higher CIR performance.

\subsection{Further Comparison with SEIZE}
\label{seize}
We observe that both  SEIZE \cite{yang2024semantic} and our STiTch generate multiple captions and apply the semantic calibration process. However, these two models are different from each other in terms of caption generation, semantic calibration strategy, and retrieval score calculation: (1) \textbf{Two-Stage Generation vs. One-Stage Generation}: SEIZE first generates $N$ captions for the reference image using a captioner and then modifies them according to the input modification text via an LLM. In contrast, our STiTch directly employs an MLLM to generate $N$ captions for the composed input, eliminating information loss from two-stage approaches. Moreover, the efficiency comparison in Tab. \ref{efficiency_vs} shows that two-stage generation methods are time-consuming, which may limit their applicability in real-time scenarios. (2) \textbf{ Similarity Space vs. Embedding Space}: SEIZE refines the final retrieval score by directly changing the cosine score. Our STiTch aims to refine the generated captions in the CLIP embedding space. (3) \textbf{Point-to-Point vs. Set-to-Set}: SEIZE represents the final global caption feature by employing the average pooling on captions, and then measures similarity with candidates via cosine similarity. Our STiTch, however, models the captions as a discrete distribution and then develops a transportation-aware set-to-set metric to calculate the distances.

\begin{figure}[htbp]
\centering
\includegraphics[width=0.48\textwidth]{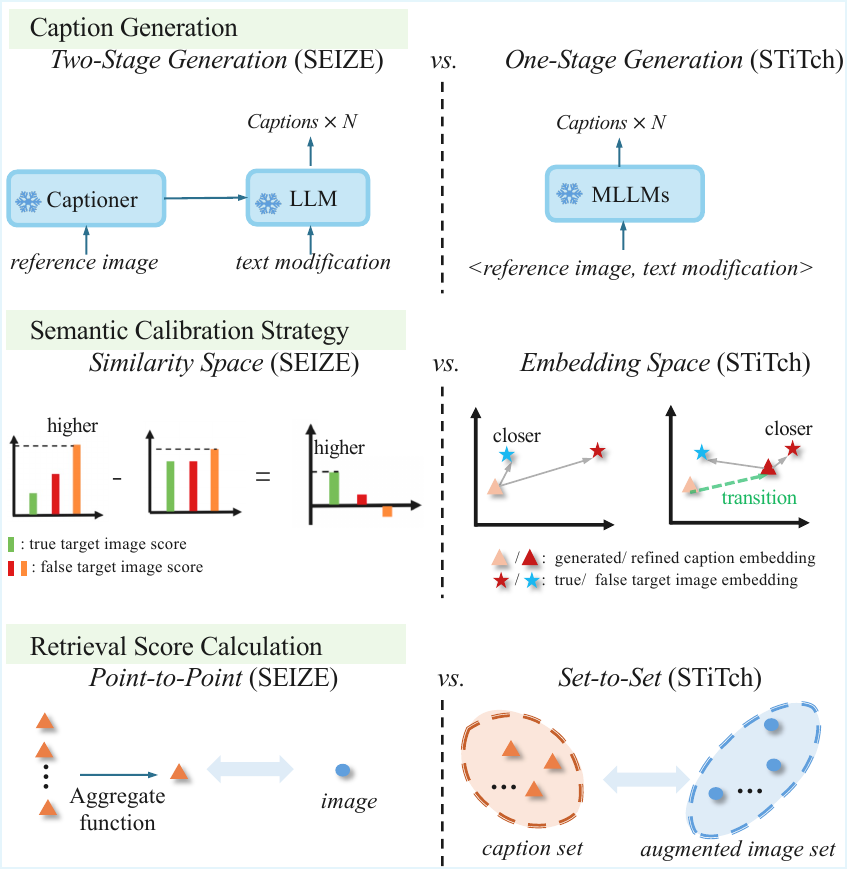}
\vspace{-3mm}
\caption{{\small{Further comparison between SEIZE and STiTch in terms of caption generation, semantic calibration strategy, and retrieval score calculation (zoom-in for more details).}}
\label{appendix_vis}}
\end{figure}

{For experiments, STiTch demonstrates more pronounced improvements on CIRR and CIRCO (Tab.\ref{appendix_c}) as well as GeneCIS (Tab.\ref{appendix_g}), where the modification descriptions are richer and require more faithful semantic modeling. 
Such settings align well with STiTch’s one-stage  caption generation and transportation-aware set-to-set metric, which jointly preserve semantic diversities and model transitions in embedding space more effectively than two-stage caption–editing pipelines.
On Fashion-IQ, however, STiTch performs below SEIZE. This is primarily due to the overly simple modification text in Fashion-IQ (\textit{e.g.}, “is solid white”, “is a lighter color”), which provides limited semantic signal and therefore offers suboptimal guidance for our transition vectors. 
SEIZE, by contrast, relies on a pre-trained captioning model to generate reference captions, a strategy more compatible with Fashion-IQ’s simplified language. However, as shown in Tab.\ref{appendix_f}, this approach incurs noticeably higher inference-time computational cost, with STiTch achieving nearly a 3× speed-up over SEIZE.}

\subsection{STiTch In-Context Learning Details}
\label{in_context}
\begin{figure*}[ht]
\centering
\includegraphics[width=0.9\textwidth]{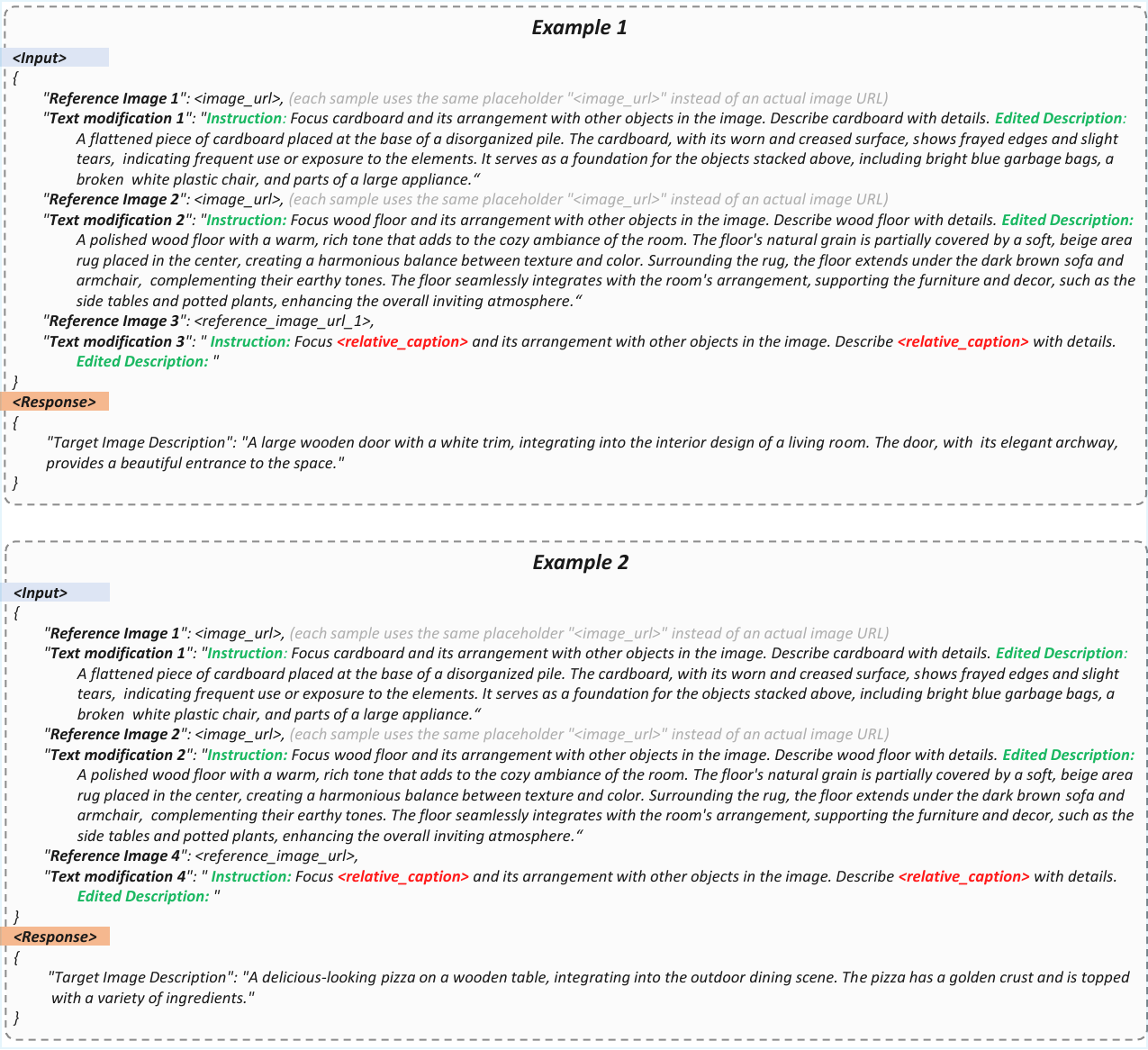} 
\vspace{-3mm}
\caption{{\small{Examples of our in-context learning on GeneCIS dataset. Each sample uses the same placeholder “\texttt{<image\_url>}” instead of an actual reference image URL.}}
\label{incontext}}.
\end{figure*}

We utilize an in-context learning method in Fig.\ref{incontext}. To achieve ZS-CIR, each sample uses the same placeholder “\texttt{<image\_url>}” instead of an actual reference image URL.
By providing several example outputs, the model is able to understand the required reasoning process without an actual reference image. This approach ensures efficient reasoning in a zero-sample setting. Each text requires the model to focus on a specific object and provide a detailed description. This helps the model understand the key elements in the image and how they relate to each other. We use uniform placeholders \texttt{<image\_url>} and \texttt{<reference\_image\_url>} to ensure that the input and output formats are consistent for easy model processing.

\begin{figure*}[htbp]
\centering
\includegraphics[width=1\textwidth]{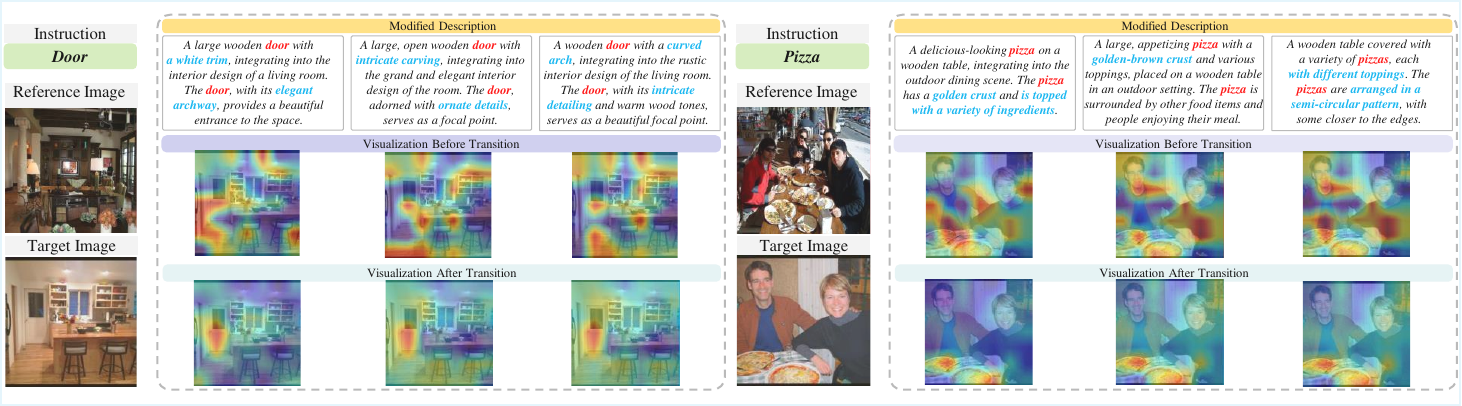} 
\vspace{-3mm}
\caption{{\small{Visualization of the GeneCIS dataset on the 'Focus Object' task. Heatmaps before and after the transition on target image are shown. Captions generated by MLLMs often contain irrelevant visual noise (\textit{\sky{blue}} text), while the STT model effectively suppresses such noise and highlights the correct focus object (\textit{\rr{red} text).
}}}
\label{appendix_vis}}
\end{figure*}

\subsection{Limitations and Future Work}
\label{future}
Although our method achieves strong performance, there remain several directions for future exploration. First, while the visual augmentation applied to target images is lightweight and performed offline, our STiTch requires approximately $M-1$ times more memory than others. We leave memory optimization as future work, with potential directions including online strategies or coarse-to-fine retrieval.
Second, when the query image depicts a complex scene involving multiple objects or relationships, and the accompanying modification text provides insufficient detail, our STiTch may focus on the wrong or ambiguous object, leading to unexpected captions. This limitation is consistent with issues observed in prior CIReVL \cite{karthik2023vision} and OSrCIR \cite{tang2025reason} models. Moreover, current benchmarks suffer from a false-negative problem. As noted in \cite{liu2021image}, each (reference image, modification) pair in FashionIQ can correspond to multiple valid target images, yet only one is annotated as ground truth. Consequently, semantically correct retrieval results may be unfairly penalized under existing evaluation protocols. We leave these challenges as promising directions for future research.

\end{document}